\documentclass[letterpaper]{article} % DO NOT CHANGE THIS
\usepackage{aaai24}  % DO NOT CHANGE THIS
\usepackage{times}  % DO NOT CHANGE THIS
\usepackage{helvet}  % DO NOT CHANGE THIS
\usepackage{courier}  % DO NOT CHANGE THIS
\usepackage[hyphens]{url}  % DO NOT CHANGE THIS
\usepackage{graphicx} % DO NOT CHANGE THIS
\usepackage{natbib}  % DO NOT CHANGE THIS AND DO NOT ADD ANY OPTIONS TO IT
\usepackage{caption} % DO NOT CHANGE THIS AND DO NOT ADD ANY OPTIONS TO IT
\usepackage{algorithm}
\usepackage{algorithmic}

\usepackage{epsfig}
\usepackage{amsmath}
\usepackage{amssymb}         
\usepackage{booktabs}       % professional-quality tables
\usepackage{amsfonts}       % blackboard math symbols
\usepackage{nicefrac}       % compact symbols for 1/2, etc.
\usepackage{microtype}      % microtypography
\usepackage[dvipsnames]{xcolor}         % colors
\usepackage{arydshln} % hdashline
\usepackage{multirow}
\usepackage{pifont}
\usepackage{bbold}
\usepackage{enumitem}
\usepackage[export]{adjustbox}
\usepackage{subcaption}
\usepackage{tablefootnote}
\usepackage{tabularx}

\usepackage{appendix}
\usepackage{my_style}
\usepackage[capitalize]{cleveref}
\crefname{section}{Sec.}{Secs.}
\Crefname{section}{Section}{Sections}
\Crefname{table}{Table}{Tables}
\crefname{table}{Tab.}{Tabs.}

\newcommand{\cmark}{\ding{51}}%
\newcommand{\xmark}{\ding{55}}

\usepackage{xspace}

\DeclareRobustCommand\onedot{\futurelet\@let@token\@onedot}
\def\onedot{. }
\def\eg{\emph{e.g}\onedot} 
\def\ie{\emph{i.e}\onedot} 
\def\cf{\emph{c.f}\onedot}

\newcommand{\datasetname}{{\fontfamily{qcr}\selectfont MultiHiEve}}

\usepackage{newfloat}
\usepackage{listings}
\DeclareCaptionStyle{ruled}{labelfont=normalfont,labelsep=colon,strut=off} % DO NOT CHANGE THIS
\floatstyle{ruled}
\newfloat{listing}{tb}{lst}{}
\floatname{listing}{Listing}
\begin{document}
% Title

% Your title must be in mixed case, not sentence case.
% That means all verbs (including short verbs like be, is, using,and go),
% nouns, adverbs, adjectives should be capitalized, including both words in hyphenated terms, while
% articles, conjunctions, and prepositions are lower case unless they
% directly follow a colon or long dash
\title{Beyond Grounding: Extracting Fine-Grained Event Hierarchies Across Modalities}
\author{
    %Authors
    % All authors must be in the same font size and format.
    % Written by AAAI Press Staff\textsuperscript{\rm 1}\thanks{With help from the AAAI Publications Committee.}\\
    % AAAI Style Contributions by Pater Patel Schneider,
    % Sunil Issar,\\
    % J. Scott Penberthy,
    % George Ferguson,
    % Hans Guesgen,
    % Francisco Cruz\equalcontrib,
    % Marc Pujol-Gonzalez\equalcontrib
    Hammad Ayyubi\textsuperscript{\rm 1},
    Christopher Thomas\textsuperscript{\rm 2},
    Lovish Chum\textsuperscript{\rm 1},
    Rahul Lokesh\textsuperscript{\rm 3},
    Long Chen\textsuperscript{\rm 4},
    Yulei Niu\textsuperscript{\rm 1},
    Xudong Lin\textsuperscript{\rm 1},
    Xuande Feng\textsuperscript{\rm 1},
    Jaywon Koo\textsuperscript{\rm 1},
    Sounak Ray\textsuperscript{\rm 1},
    Shih-Fu Chang\textsuperscript{\rm 1}
}
\affiliations{
    %Afiliations
    \textsuperscript{\rm 1}Columbia University,
    \textsuperscript{\rm 2}Virginia Tech,
    \textsuperscript{\rm 3}Samsung Research America,
    \textsuperscript{\rm 4}HKUST \\
    
    % If you have multiple authors and multiple affiliations
    % use superscripts in text and roman font to identify them.
    % For example,

    % Sunil Issar\textsuperscript{\rm 2},
    % J. Scott Penberthy\textsuperscript{\rm 3},
    % George Ferguson\textsuperscript{\rm 4},
    % Hans Guesgen\textsuperscript{\rm 5}
    % Note that the comma should be placed after the superscript

    % email address must be in roman text type, not monospace or sans serif
    hayyubi@cs.columbia.edu
%
% See more examples next
}

%Example, Single Author, ->> remove \iffalse,\fi and place them surrounding AAAI title to use it
\iffalse
\title{My Publication Title --- Single Author}
\author {
    Author Name
}
\affiliations{
    Affiliation\\
    Affiliation Line 2\\
    name@example.com
}
\fi

\iffalse
%Example, Multiple Authors, ->> remove \iffalse,\fi and place them surrounding AAAI title to use it
\title{My Publication Title --- Multiple Authors}
\author {
    % Authors
    First Author Name\textsuperscript{\rm 1},
    Second Author Name\textsuperscript{\rm 2},
    Third Author Name\textsuperscript{\rm 1}
}
\affiliations {
    % Affiliations
    \textsuperscript{\rm 1}Affiliation 1\\
    \textsuperscript{\rm 2}Affiliation 2\\
    firstAuthor@affiliation1.com, secondAuthor@affilation2.com, thirdAuthor@affiliation1.com
}
\fi

% REMOVE THIS: bibentry
% This is only needed to show inline citations in the guidelines document. You should not need it and can safely delete it.
% \usepackage{bibentry}
% END REMOVE bibentry

% \maketitle

\maketitle
\thispagestyle{empty}

\startconceptfig

{
{
\begin{center}
\resizebox{0.9\linewidth}{!}{
\includegraphics[width=\textwidth]{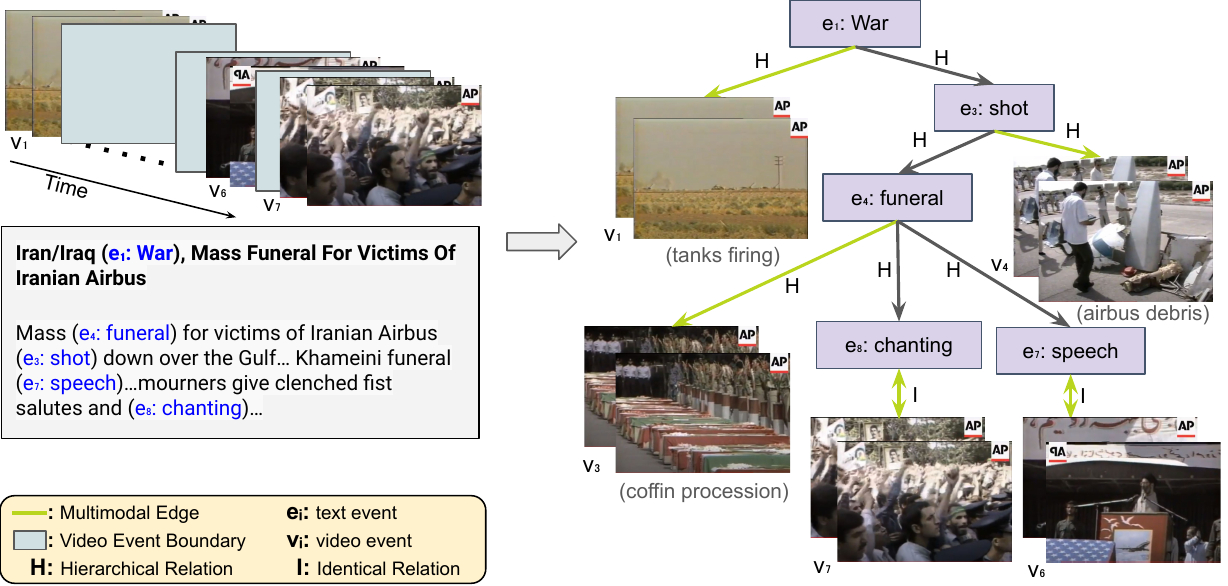}
}
\end{center}
}

\begin{minipage}{6.75in}
\captionof{figure}{An example from our {\fontfamily{qcr}\selectfont MultiHiEve} dataset illustrating the proposed task of extracting event hierarchies given multimodal (text + video) content. The same legend is followed throughout the paper in all illustrations of event hierarchies.}
\label{fig:intro_war}
\end{minipage}
\vspace{0.5em}
}

\endconceptfig

\begin{abstract}
% Understanding how events described or shown in multimedia content relate to one another is a critical component to developing robust artificially intelligent systems which can reason about real-world media. While much research has been devoted to event understanding in the text, image, and video domains, none have explored the complex relations that events experience \textit{across} domains. For example, a news article may describe a `protest' event while a video shows an `arrest' event. Recognizing that the visual `arrest' event is a subevent of the broader `protest' event is a challenging, yet important problem that prior work has not explored. 
% In this paper, we propose the novel task of \textbf{M}ulti\textbf{M}odal \textbf{E}vent \textbf{E}vent \textbf{R}elations ($\textrm{M}^{2}\textrm{E}^{2}\textrm{R}$) to recognize such cross-modal event relations. To assist research on this task, we contribute a large-scale dataset consisting of 100k video-news article pairs, as well as a benchmark of densely annotated data. We also propose a weakly supervised multimodal method to predict rich multimodal event relations. Experiments show that our model outperforms a number of competitive baselines on our proposed benchmark. We also perform a detailed analysis of our model's performance and suggest directions for future research.

Events describe happenings in our world that are of importance. %Naturally, 
%Hammad: the sentences aren't flowing otherwise
Naturally, understanding events mentioned in multimedia content and how they are related forms an important way of comprehending our world.
Existing literature can infer if events across textual and visual (video) domains are identical (via grounding) and thus, on the same semantic level.
% However, event-event relations are far more complex as events are referred to on many semantic/granular levels across domains.
% For example, in \Cref{fig:intro_war}, airplane ‘shot’ event in text and ‘tanks firing’ event in the video are ‘part-of’ of the same main event ‘war’ and refer to it in a fine-grained manner. Hence, airplane ‘shot’ and ‘tanks firing’ are hierarchically related to ‘war’.
However, grounding fails to capture the intricate cross-event relations that exist due to the same events being referred to on many semantic levels.
For example, in \Cref{fig:intro_war}, the abstract event of ``war'' manifests at a lower semantic level through subevents ``tanks firing'' (in video) and airplane ``shot'' (in text), leading to a hierarchical, multimodal relationship between the events.
% Together, they form a hierarchy of events. 
% Resolving these semantic granularities via event hierarchies reveal the structure of events and is key to understanding them.
%We propose to 

% Much of the prior work on extracting such event hierarchies has been done in NLP for text-only domain. However, as our world is multi-modal, the information conveyed by unimodal text event hierarchy is inherently limited. % For example, in figure 1, ‘burning’ subevent from video happening during parent event ‘protest’ point to the fact that the protest was not peaceful.
%As such, we propose the task of extracting event hierarchies given multimodal data – text article and paired video. 
In this paper, we propose the task of extracting event hierarchies from multimodal (video and text) data to capture how the same event manifests itself in different modalities at different semantic levels.
This reveals the structure of events and is critical to understanding them.
% in a fine-grained structure which we automatically extract from real-world media.
To support research on this task, we introduce the \textit{\textbf{Multi}modal \textbf{Hi}erarchical \textbf{Eve}nts} {\fontfamily{qcr}\selectfont(MultiHiEve)} dataset\footnote{\url{https://github.com/hayyubi/multihieve}}. 
Unlike prior video-language datasets, {\fontfamily{qcr}\selectfont MultiHiEve} is composed of news video-article pairs, which makes it rich in event hierarchies.
We densely annotate a part of the dataset 
% with multimodal event hierarchies 
to construct the test benchmark.
We show the limitations of state-of-the-art unimodal and multimodal baselines on this task.
Further, we address these limitations via a new weakly supervised model, leveraging only unannotated video-article pairs from {\fontfamily{qcr}\selectfont MultiHiEve}. 
%Despite these advances, the task remains challenging and far from being solved. 
%We finally provide an analysis of some of these challenges and suggest avenues for future work.
We perform a thorough evaluation of our proposed method which demonstrates improved performance on this task and highlight opportunities for future research.
\end{abstract}

\begin{figure}
\begin{minipage}[t][239pt][t]{\textwidth}
\end{minipage}
\end{figure}

\section{Introduction}

Human life is eventful. We use events to describe what is happening (e.g.~war, protest, etc.), to tell stories (e.g.~during the war an airplane was shot down), and to depict our understanding of the world (e.g.~coffin procession happens in a funeral). %Understanding the connection between events %and their relations
Thus, understanding and analyzing events is a crucial part of comprehending our world.
A critical component towards this goal is to figure out the manner in which the same real-world event is manifested in multiple modalities of data.
% is thus a crucial step towards the goal of understanding these events and their different components.

% Towards this goal, existing work can predict if events across textual and visual domains are on the same semantic level and are thus related identically, via the task of grounding \cite{gao2017tall}. However, events across domains are referred to on many semantic levels, forming complex hierarchical and sibling relations. 
% Whenever we talk about a (parent) event, inevitably, it comprises of certain (sub) events, leading to a hierarchical relationship between the parent event and the subevents, and sibling relationships among the subevents. 
% For example, we see from \Cref{fig:intro_war} that event `tanks firing' shows a component of `war' and refers to it on a finer semantic level. Hence, `tanks firing' holds a hierarchical subevent-parent event relationship with `war'. 
% In addition, `tanks firing' event in video and airplane `shot' event in text together reveal what comprises `war' event, and are thus siblings. 
% `war'event is comprised of `tanks firing' and airplane `shot' events and is thus hierarchically related to them. 
% Further, the subevents could be composed of sub-subevents, creating an event hierarchy (\cf  \Cref{fig:intro_war}). 
% These event hierarchies relate events according to the semantic scale at which they occur, revealing a hierarchical compositional structure. This makes them critical to understanding events and their fine-grained relationships.

To this end, previous studies have utilized %the task of 
grounding \cite{gao2017tall} to determine whether events in textual and visual domains are related identically at the same semantic level. 
However, events in different domains can be referred to at various semantic levels, resulting in intricate hierarchical and sibling relationships.
For instance, as illustrated in \Cref{fig:intro_war}, the event of ``tanks firing'' is a component of event ``war'' and denotes it at a finer semantic level. 
Consequently, ``tanks firing'' is a subevent of the parent event ``war''. 
Moreover, textual event of ``airplane shot'' is also a subevent of ``war'', and together with ``tanks firing'' reveal the constituents of ``war'' event. This creates siblings relations between ``airplane shot'' and ``tank firing''.
Additionally, subevents can be further decomposed into sub-subevents, creating a hierarchy of events (see \Cref{fig:intro_war}). 
These event hierarchies organize events based on the semantic scale at which they occur and expose a hierarchical compositional structure, which is crucial for understanding events and their fine-grained relationships. 
% Note that since grounding grounds events at the same semantic level, grounding-based methods never produce event hierarchies.

% \chris{Prior work \cite{} in the natural language processing community has proposed capturing event relationships in a graph structure called an ``event hierarchy'' which describes how events are related to one another. The event hierarchy captures the hierarchical compositional structure of complex events, making it an important tool in understanding events and their fine-grained relationships. This structured representation has found use in many downstream tasks requiring eventful reasoning: summarization \cite{}, question answering \cite{}, and commonsense reasoning \cite{}. (can we cite NLP papers here that use event graphs for these tasks?}

% \chris{One limitation of prior text-only event understanding methods is that events which relate to one another exist in different modalities. Understanding cross-modal event relationships remains a largely unaddressed, yet highly important, problem. Existing work \cite{} can predict, to a certain extent, if events in text and in visual content are the same via grounding. However, how such events are related is often much more nuanced. For example, the Russian invasion of Ukraine event was itself comprised of many subevents such as tank and troop movements and deaths, which are distinct from the war itself. However, existing grounding methods are limited in that they only detect grounding between image and text but are unable to differentiate fine-grained event relationships. }

Much of the prior work on extracting such event hierarchies has been done in Natural Language Processing (NLP) for the text-only domain. However, as our world is multi-modal, the information conveyed by a unimodal text event hierarchy is inherently limited. For example, in \Cref{fig:task_diff}, extracting ``evacuation'' subevent from video as a child of parent event ``fire'' provides us with the additional fact that relief efforts reached on time.

% \chris{In this work, we are the first to address predicting cross-modal event relationships. We propose the novel task of extracting event hierarchies from multimodal (text and video) data. Given events from video-article pairs, the task requires understanding how events across modalities relate to each other. Specifically, this involves distinguishing between those events which directly appear visually (\ie ~an identical cross-modal relation) and those which are related at different semantic levels (\ie ~hierarchical).
% This output can be combined with text-only event hierarchy (from any off-the-shelf tool) to get a more holistic hierarchy. 
% }

We address these limitations through the proposed task of extracting event hierarchies from multimodal (text \& video) data. Specifically, given events from paired text article and video, the task requires predicting all the multimodal hierarchical and identical event-event relationships. This output can be combined with text-only event hierarchy (from any off-the-shelf tool) to get a more holistic hierarchy. 

% A multimodal event hierarchicy can aid many downstream tasks: summarization (from \Cref{fig:intro_protest}, we can summarize that marching, burning tires and tear gas firing happened during protest), question answering (from \Cref{fig:intro_protest}, the question `What are so many people doing on the street?' can be answered as protesting because the frame containing people on streets is hierarchically related to protest event), commonsense reasoning (from \Cref{fig:intro_protest}, we can reason that the protest was not peaceful because it contained burning as a subevent), visual entailment (from \Cref{fig:intro_protest} we can say that people in the jeep are not going for picnic because the frame of jeep is identically related to firing a tear gas) etc.
Multimodal event hierarchies can aid many applications, such as summarization \cite{mg-text-summarization}, story completion from multiple sources, event analysis/comparison (\eg a ``protest'' event with ``property destruction'' subevent is unruly, otherwise it's peaceful), event prediction likelihood \cite{mg-next-event}, knowledge-based information extraction \cite{wen-etal-2021-resin}, and multimodal knowledge graph construction \cite{li-etal-2020-gaia}.

To study this task, we introduce the \textit{\textbf{Multi}modal \textbf{Hi}erarchical \textbf{Eve}nts {\fontfamily{qcr}\selectfont(MultiHiEve)}} dataset. {\fontfamily{qcr}\selectfont MultiHiEve} consists of approximately 100.5K pairs of news article and accompanying video. The news story in the text article mentions events on multiple semantic levels, making it ideal for the task of extracting event hierarchies.  We strive to limit the socio-economic bias inherent in news media by only collecting our data from news sources rated unbiased by credible sources. We keep unannotated 100K pairs for training and densely annotate 526 pairs with multimodal hierarchical and identical relations for benchmarking and evaluation. 
Our annotation process is detailed and labor-intensive, requiring approximately 114 hours of expert annotator effort. Crucially, in contrast to prior text-only datasets dealing with hierarchical events, we do not limit the event types to any fixed ontology and instead consider an open world of events.

To benchmark performance on this task, we construct several baselines using state-of-the-art (SOTA) architectural components. A unimodal text-only baseline leverages ASR (automated speech recognition) and employs a SOTA NLP model \cite{wang-etal-2021-learning-constraints} to find hierarchical events between a text article and its video's ASR. We also build a multimodal baseline by detecting hierarchical events in text using \citet{wang-etal-2021-learning-constraints} and grounding the textual subevents to video using CLIP \cite{clip-2021}. 
A key limitation of these baselines is that they require visual subevents to be mentioned in textual form in either the ASR or the article. To address this, we propose \textbf{M}ultimodal \textbf{A}nalysis of \textbf{S}tructured \textbf{H}ierarchical \textbf{E}vents \textbf{R}elations (MASHER), a weakly supervised model which learns to directly predict hierarchical events between text and video. By doing so, MASHER can also discover visual-only subevents (subevents not mentioned in text).

%MASHER acquires this ability by training on multimodal pseudo labels generated by the aforementioned multimodal baseline.

The major contributions of this work are fourfold: 1) We propose the challenging task of extracting event hierarchies from multimodal data. 2) We release {\fontfamily{qcr}\selectfont MultiHiEve} dataset to facilitate research on this task. 3) We construct several baselines and propose MASHER, a weakly supervised model, to benchmark performance. 4) We provide a detailed analysis of our dataset and methods with insights for future work.

% The major contributions of this work are fourfold: 1) We propose the challenging task of extracting event hierarchies from multimodal data, and 2) release a dataset, {\fontfamily{qcr}\selectfont MultiHiEve}, to facilitate research on this task. 3) We do not restrict the type of events to any fixed ontology. 4) We construct several baselines and propose MASHER, a weakly supervised model, to benchmark performance on this task.

\section{Related Work}
\label{sec:related_work}
% \begin{itemize}
%     \item Hierarchical Event Relations in Text
%     \item Hierarchical relations in video (vid situ)
%     \item Video + Language Tasks
%     \item M2E2 (Alireza and Xudong), visual comet, next event Mohit's group, Manling events, UPenn work, Grounding (Coreferencing)
%     \item Scene graph Generation (from video/images) and its difference with event graphs
%     \item Explain how your task is different from these other ones.
% \end{itemize}
%-------------------------------------------------------------------------
\textbf{Hierarchical Event Relations in Text.} Detecting hierarchical event relations (or sub-event relations) is a long-standing problem in the text domain~\cite{o2016richer,glavavs2014hieve}. % By ``hierarchical", we mean that the parent and child event are spatiotemporal containment. 
Early works mainly rely on heuristic phrasal patterns. For example, \citet{badgett2016extracting} found some characteristic phrases (\eg ``media reports'' in new articles) always contain sub-events with hierarchical relations. To further enrich hierarchical event relation instances, recent works~\cite{yao2020weakly} rely on generative language model to generate subevent knowledge among different commonsense knowledge~\cite{bosselut2019comet,sap2019atomic}, then incorporate knowledge into event ontology. 
% In contrast, we focus on multimodal hierarchical event relations. 

\begin{figure}[t]
    \centering
    \includegraphics[width=\linewidth]{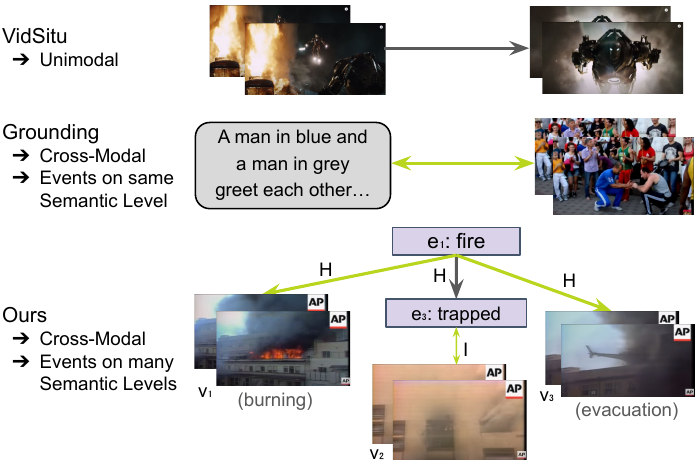}
    \caption{
    % An example from {
    % \fontfamily{qcr}\selectfont MultiHiEve} showing visual-only subevent, ``burning tires'' ({\color{orange}orange bounded}), which can't be discovered via prior NLP methods.
    % Significance of the proposed event hierarchy extraction task. The {\color{orange}orange bounded} subevent (burning tires) is a visual-only subevent, which can't be discovered via prior NLP methods. The bottom row shows downstream applications which can be driven by a mutltimodal event hierarchy. VQA: Visual Question Answering. VE: Visual Entailment. The {\color{Cerulean}blue box} shows the task and {\color{YellowGreen}green box} contains the answer. The color of the task name co-relates with the color of the edge used to solve that task.
    Illustration of our task's differences with other related tasks. Unlike prior tasks, our task is multimodal and relates events on multiple semantic scales.
    }
    \label{fig:task_diff}
\end{figure}

\textbf{Relation Understanding in the Vision Domain.} Prior work \cite{krishna2017visual,xu2017scene,ji2020action} propose scene graph methods that parse images/videos into a graph. %, where nodes are usually visual entities and edges are the relationships between them. 
However, the relationships studied in scene graphs are not between two events.
%\noindent \textbf{Event Relation in Video Understanding.}
To the best of our knowledge, the only pioneering work that has discussed the event-event relationship in video domain is VidSitu~\cite{Sadhu_2021_CVPR}. Unfortunately, to simplify the research problem on this topic, they have made several assumptions: 1) All events are manually cutted into fixed interval (2-second). 2) All event types are ``visual'' only and from a fixed event ontology. On the other hand, we consider variable-length video events and focus on open-vocabulary event types (which include non-visual event types and other visual events like funeral, detain, rally etc). Besides, while they have annotated each video event with a text label, their event relations still are between events in a video. In contrast, our multimodal relations are between events in an article and a video (see \Cref{fig:task_diff}).
%Additionally, the hierarchical event relationship is much more common than the causal relationships in VidSitu.

% \noindent \textbf{General Video$+$Language Tasks.}
% Many datasets and benchmarks have been proposed during the past decade for video$+$language tasks, such as video question answering~\cite{Tgif-qa, MovieFIB, MovieQA, lei2018tvqa}, video captioning~\cite{MSRVTTXuEtal:CVPR2016, KrishnaEtAl:ICCV2017, YouCook2ZhouEtAl:AAAI2018, lei2020tvr}, audio-visual dialog~\cite{AVSD}, video entailment~\cite{liu2020violin}, future event prediction~\cite{lei2020more} and etc.. However, to the best of our knowledge, none of them studied the problem of understanding relationships between a textual event and a video event.

\textbf{Multimodal Event Understanding.}
Since single-modality event tasks are well studied~\cite{nguyen-etal-2016-joint,sha2018jointly,liu-etal-2019-neural-cross,liu2020event,li2017situation,mallya2017recurrent,pratt2020grounded, Yatskar2016SituationRV, Lu_2023_CVPR}, jointly understanding events from multiple modalities~\cite{li2020cross,chen2021joint,li2022clip,zhang2017improving,tong2020image,wen-etal-2021-resin,park2020visualcomet,reddy2022mumuqa,du2022resin} has attracted extensive research interests because different modalities usually provide the complementary information for comprehensively understanding the real-world complex events. Two important benchmarks~\cite{li2020cross,chen2021joint} have been established for \textit{image + text} and \textit{video + text} settings. \citet{li2020cross} first introduced the task of jointly extracting events and labeling argument roles from both text articles and images. 
%An unannotated image-article dataset and a manually labeled evaluation image-article dataset are collected to train and evaluate models, respectively. 
~\citet{chen2021joint} further defined the task of joint multimedia event extraction from video and text to exploit the rich dynamics from videos. % Both of these two benchmarks~\cite{li2020cross,chen2021joint} also explored coreference between visual and textual events. 
% However, neither of them handle the fact that visual events are usually low-level (e.g., punch) compared to textual events (e.g., attack) and further study the fine-grained relationship, e.g., hierarchical.
However, both the works focus on event detection in comparison to the event relations task explored in this work.

% \begin{figure}[t]
%     \centering
%     \includegraphics[width=\linewidth]{images/difference_with_grounding3.pdf}
%     \caption{An example from {\fontfamily{qcr}\selectfont MultiHiEve} showing difference with grounding task: grounding can't relate subevent ``evacuation'' to ``fire'' event as they are not visually similar.}
%     \label{fig:diff_with_grounding}
% \end{figure}

\section{Task}
% Hammad: I don't want to mention the word related because that would mean dealing with more relation types and reviewers questions relating to same. Also, don't want to mention downstream applications at this time, which can again lead to reviewers concerns.
% Our proposed task requires predicting how events expressed in different modalities are related. Knowing these fine-grained event relationships allows one to construct a multimodal event hierarchy which captures the compositional structure of events and which can facilitate downstream reasoning.

To understand (parent) events and fully comprehend what they entail, one needs to discover what (sub) events happened during the parent event. The task of extracting event hierarchy from multimodal content is aimed at revealing this compositional structure of events.

\textbf{Formal Task Definition.}
\label{sec:task}
Given a text article, $T$, containing events, $\{e_i\}_{i=1}^m$, and 
a  video, $V$ containing events $\{v_j\}_{j=1}^n$, the proposed task requires prediction of all possible hierarchical and identical event-event relations, $\{r_k\}_{k=1}^K$, from a text event, $e_i$, to a video event, $v_j$, 
among all possible $m \times n$ pairs, where $r_k \in \{\textrm{'Hierarchical', 'Identical'}\} \textrm{ and } K \leq m \times n$. 
We will now discuss definitions for different components of the task and justifications for task design choices.

\textbf{Text Event Definition.} The definition of an ``event'' has been defined quite thoroughly in different NLP works on information extraction \cite{acecorpus2014, timebank2016, graphevents2015}. 
As such, we closely follow ACE Corpus's \cite{acecorpus2014} definition of an event:
% Very broadly it's defined as: 
`\textit{a change of state or the changed state of a person, thing or entity.}'
% ACE also provides a more detailed criteria addressing linguistic nuances not required in our setting. 
% Secondly, \cite{acecorpus2014} is restricted to a fixed ontology of events, whereas we annotate events in an open world. 
% As a result, we came up with a modified event definition and annotation criteria (detailed in 
%Appendix)
% \Cref{appendix:text_event_def}).
We came up with a slightly modified event definition and annotation criteria (detailed in 
%Appendix)
\Cref{appendix:text_event_def}) as ACE 1) addresses several linguistic nuances not required in our setting;
2) is restricted to a fixed ontology of events, whereas we annotate open-domain events.

\textbf{Video Event Definition.} Precisely defining what constitutes an ``event'' in the video domain is challenging due to the multiple granularities at which events occur in videos.
For example, during a \textit{``clash''} event, one might see a \textit{``pulling} out baton'' event and a \textit{``throwing} a punch'' event. 
This makes it difficult to pick salient event boundaries in video clips.
\citet{Sadhu_2021_CVPR} circumvent this ambiguity by defining temporal event boundaries of fixed duration (two seconds).
However, pre-defining the boundary duration is difficult and application specific. Additionally, a fixed duration boundary often divides salient events into multiple segments. 
We address these issues by defining video event boundaries to be where shot changes occur, partly following \cite{genericEventBoundary2021}. From our qualitative analysis and annotator feedback, this gives us a good trade-off between ease, clarity, consistency and non-segmentation of events.
% based on qualitative analysis and annotators feedback.

\textbf{Relation Types.}
We define two types of event relations in this work: hierarchical and identical. These relation types are well defined in NLP \cite{glavavs2014hieve} and we follow them to define the relations for our task as below:

Hierarchical:  \textit{``A parent event $A$ is said to be hierarchically related to a subevent $B$, if event $B$ is spatio-temporally contained within the event $A$.''} For example, an ``evacuation'' event is a subevent of a ``fire'' event as it takes place
during and at the same location as the fire event (see \cref{fig:task_diff}). Therefore, a subevent (evacuation) is a component of the parent event (fire) among multiple other subevents (burning, trapped, evacuation etc.).

Identical: \textit{``An event $A$ is said to be identical to another event $B$ if both events denote exactly the same real-world events in all aspects.''} For example, ``trapped'' event in text is identical to the video event showing people begin trapped as they both denote exactly the same event -- there are no more components of trapped.

\textbf{Relation direction.} The multimodal relations in our event hierarchies are directed from text event to video events. The logic behind this design choice is that text events are often more abstract while video events are often atomic. For example, we are likely to observe abstract events such as war and election in text while their atomic subevents -- fighting and voting -- are more likely to be visible in the video.

\textbf{Difference from Video Grounding. }
Although grounding relates similar events in text and video, it does not distinguish the type of relationship. That is, whether the video event shows all aspects of text event (\ie ~identical) or whether it only shows ``part-of'' of the parent event and is thus a subevent. 
This has major implications. 
For example, in \Cref{fig:task_diff}, inferring that ``burning'' video subevent is identical to ``fire'' would imply that there was nothing else that happened during fire event and hence, relief efforts did not reach on time. On the other hand, inferring ``burning'' to be a subevent of ``fire'' indicates that there may have occured other relief/``evacuation'' subevents.
Further, some video subevents are visually dissimilar to its textual parent event (for example, ``evacuation'' is dissimilar to ``fire'' in \Cref{fig:task_diff}), making it difficult for grounding to relate such subevents.  

% An argument could be made that since subevents look visually similar to the parent event, these subevents could be discovered using video grounding. The fallacy in the argument is that not all subevents look visually like their parent event. For example, in \cref{fig:diff_with_grounding}, subevent `evacuation' doesn’t have much visual similarity to its parent event `protest'. More importantly, although grounding relates similar events in text and video, it doesn’t distinguish if the video event shows all aspects of text event and is thus identical or it only shows a `part-of’ of the parent event and is thus a subevent. 
% This has major implications. 
% For example, \Cref{fig:intro_protest}, inferring that `marching' video subevent is identical to `protest' would imply that there was nothing else that happened during protest and hence, it was peaceful. On the other hand, inferring `marching' to be a subevent of `protest' would imply that protest may or may not have contained violent subevents.

%As such, it will be difficult for a video grounding model to predict these `Hierarchical' relations.

%-------------------------------------------------------------------------
\begin{table}[t]
\centering
\small
% \resizebox{!}{\textwidth}{
\setlength{\tabcolsep}{2pt}
\scalebox{0.8}{
\begin{tabular}{l l l l}
\toprule
Dataset       & Domain  & Text Type & Words/min.\tablefootnote{For datasets with duplicate descriptions per video clip (MSVD, MSR-VTT), words/min. is averaged by \#descriptions .} \\
\midrule
MSVD~\cite{chen2011collecting}         & Open    & Caption         &          54     \\
% MPII Cooking~\cite{rohrbach2012database}  & Cooking & 44                            & 600                                      & 8                                  & No  & 5609                             \\
% TACoS~\cite{regneri2013grounding}         & Cooking & 127                           & 360                                      & 15.9                               & No  & 18K                              \\
% TACos-MLevel~\cite{rohrbach2014coherent}  & Cooking & 185                           & 360                                      & 27.1                               & No  & 53K                              \\
% M-VAD~\cite{torabi2015using}         & Movie   & 92                            & 6.2                                      & 84.6                               & No  & 56K                              \\
MSR-VTT~\cite{xu2016msr}       & Open    & Caption  & 38 \\
Charades~\cite{sigurdsson2016hollywood}      & Daily activities   &   Caption  & - \\
% VTW~\cite{zeng2016generation}           & Open    & 18.1K                         & 90                                       & 213.2                              & No  & 45K                              \\
% YouCook II~\cite{ZhXuCoAAAI18}    & Cooking & 2000                          & 316                                      & 176                                & No  & 15K                              \\
ActyNet Cap~\cite{krishna2017dense}   & Open    & Caption & 27\\
HowTo100M~\cite{miech2019howto100m} & Instructional & ASR  & 67\\
YT-Temporal-180M~\cite{merlot-2021} & Open & ASR & -\\
\color{gray} VidSitu~\cite{Sadhu_2021_CVPR} & \color{gray}Movie & \color{gray}Null & \color{gray}Null \\
\midrule
{\fontfamily{qcr}\selectfont \textbf{MultiHiEve}} & News    & News Story & 113 \\ 
\bottomrule
\end{tabular}
}
\caption{Comparing {\fontfamily{qcr}\selectfont MultiHiEve} to prior video-language datasets. It is the first dataset sourced from news domain and contains a story in the text description. Grayed row denotes video-only dataset.}
\label{tab:train-dataset-modified}
\end{table}
\section{Dataset}
\label{sec:dataset}

To support research on the proposed task, we introduce {\fontfamily{qcr}\selectfont MultiHiEve} -- a dataset containing news articles and the associated video clips. 
Existing video-language datasets contain either manually annotated descriptions of video events or utterances from the video itself (see \Cref{tab:train-dataset-modified}). In both cases, the text is essentially on the same semantic level as the video event. However, they lack a context or an overall story describing events on higher semantic levels that comprise the video events. In contrast, news stories provide a rich 
% multimodal (\eg  images and videos)
hierarchy of events, making them ideal for our task. Having $\sim$~2x more words per minute as compared to other datasets (see \Cref{tab:train-dataset-modified}), indicates this to some extent.

% As such, we collect our dataset from news domain which provides the ideal testbed for extracting . We compare our dataset to popular video
% as the content is rich in event mentions and their relations \cite{glavavs2014hieve, hovy-etal-2013-events}. 

\subsection{Data Collection and Curation}
A potential drawback with news data is that they could be socio-economically biased and sensationalized. We mitigate these issues by choosing media sources rated ``Center'' (out of ``Left'', ``Left Leaning'', ``Center'', ``Right Leaning'' and ``Right'' ratings) by the media rating website \texttt{allsides.com}, resulting in a total of 9 news media sources 
%(\cf  Appendix).
(\cf \Cref{appendix:dataset-curation}). 
We scraped Youtube for news videos, associated text story and closed captions (ASR) from the official channels of these sources, collecting a total of 100.5K videos. We filtered videos whose duration was greater than 14 minutes or whose descriptions had less than 10 words. This was done to prune videos that may be too computationally expensive to process or whose descriptions may be too short to have meaningful events. We split the data two ways -- 1) 100K unannotated train split for self-supervised/weakly supervised training and 2) 526 annotated test split - 249 validation set and 277 test set - for benchmarking and evaluation. We annotate a relatively small set because of the challenging and resource-consuming nature of the annotation process; two popular NLP Hierarchical event-event relations datasets \cite{hovy-etal-2013-events, wang-etal-2021-learning-constraints} contain 100 articles each (including train split).

\begin{table}[t]
\centering

\small
\setlength{\tabcolsep}{2pt}
\scalebox{0.9}{
\begin{tabular}{lllll}
\toprule
Dataset      & Domain & Modality & \#Hier. Rels. & \#Id. Rels. \\
\midrule
HiEve~\cite{glavavs2014hieve}        & News   & Text          & 3648                          & 758                         \\
IC~\cite{hovy-etal-2013-events}          & News   & Text          & 4586                          & 2353                        \\
\midrule
{\fontfamily{qcr}\selectfont MultiHiEve} & News   & Text + Vision & 3077                          & 1524  \\
\bottomrule
\end{tabular}
}
% \caption{Comparison of {\fontfamily{qcr}\selectfont MultiHiEve} test set against other datasets containing hierarchical events. ``Hier.'' denotes ``Hierarchical''. ``Id.'' denotes ``Identical''.}
\caption{Comparison of {\fontfamily{qcr}\selectfont MultiHiEve} against other hierarchical-event datasets. ``Hier.'' denotes ``Hierarchical''. ``Id.'' denotes ``Identical''.}
\label{tab:test-dataset}
\end{table}
\subsection{Train Split}

The train split contains 100K videos with a total duration of more than 4K hours. The paired text descriptions total 1.9M sentences and 28M words. The large-scale nature of the data allows for self/weakly-supervised learning on the task. 
% We additionally provide a quantitative comparison of our train split against 12 popular video language datasets in \Cref{tab:train-dataset}. 
%(more analysis in Appendix).
% \Cref{appendix:dataset}). 
% Our dataset is orders of magnitude larger than comparable datasets, both in terms of video duration and text article length ($\sim$20x), which provides a fertile source of events. Additionally, our dataset is the only one that includes Automatic Speech Recognition (ASR) transcripts along with article descriptions of videos. 
% We compare our dataset against more video-language datasets, provide additional data statistics and explore the dataset's topic distribution in \Cref{appendix:dataset-exploration}.
We provide additional data statistics, topic distribution exploration and quantitative comparison against 12 popular video-language datasets in \Cref{appendix:dataset-exploration}.

\subsection{Test Split}
\label{sec:dataset-test}

\textbf{Annotation Procedure.} As a first step, following the definition of a video event from \Cref{sec:task}, we extract video events using an off-the-shelf video segmentation model: PySceneDetect \footref{footnote:pyscene}. To make text event annotation easier, we provide automatically extracted text events (using \cite{shen-etal-2021-corpus}) to the annotators along with instructions to add or omit events according to the definition in \Cref{sec:task}. Next, we task the annotators to mark all possible relations, $\in$ {``Hierarchical'', ``Identical''}, from the annotated text events to the provided video events in a video-article pair. We provide screenshots of our annotation tool and additional details in 
%Appendix. 
\Cref{appendix:dataset-annotation}.

We train 5 expert annotators for this task through a series of short seminars and multiple rounds of feedback and consultation with all the annotators to improve consensus. Excluding training, annotation required 114 hours in total, reflecting the labor-intensive and complex nature of the task.

% \vspace{3pt}
\textbf{Inter Annotator Agreement (IAA).} We measure the quality of the annotations using IAA. Inspired by \cite{glavavs2014hieve} and \cite{graphevents2015}, we formulate IAA$_j = \frac{\sum_r\mathbb{1}(x_{rj}\geq 2)}{|\cup_{i=1}^5S_j^i|}$, where $j\in$\{``Hierarchical'', ``Identical''\}, $S^i_j$ denotes the set of all relations annotated by annotator $i$ as $j$, $r \in \cup_{i=1}^5S^i_j$ and $x_{rj}$ represents \# annotators who marked relation $r$ as $j$. 
The intuition behind this formulation is to calculate the percentage of relations which have been annotated by at least 2 annotators. We obtain IAA$_{Hierarchical}=47.5$ and IAA$_{Identical}=48.9$. 
%While the IAA score may seem low, 
This is not far from IAA$_{Hierarchical}$ for text datasets HiEve and IC -- 69 and 62 respectively. Thus, while the text-only event relation task is itself quite challenging, our new cross-modal task is even more demanding.

Following prior work for related text-only tasks \cite{vulic2019multilingual, glavavs2014hieve}, we consider IAA to be an upper bound on model performance because our metrics judge the model's predictions with respect to human agreement on the task. It is not clear whether a model exceeding IAA indicates a meaningful performance gain or an overfitting to annotators' subjective tendencies.
%We note that whether IAA should be considered an upper-bound on model performance is debated \cite{richie2022inter}.
% , lowering the IAA score even further.

\textbf{Dataset Analysis.} As we are the first to propose \textit{multimodal} hierarchical event relations analysis, we compare our dataset against two popular \textit{text-only} hierarchical event relations datasets in \Cref{tab:test-dataset}. Overall, {\fontfamily{qcr}\selectfont MultiHiEve} has a comparable number of hierarchical and identical relations, but has the added novelty of being the first multimodal (text and vision) event-event relations dataset. Further, both NLP datasets limit the event types to a fixed ontology. We do not put any such constraints on either text or video event types.
%-------------------------------------------------------------------------

\section{Multimodal Hierarchical Events Detection}

Acquiring a large scale labelled dataset sufficient for training a model on the proposed task is prohibitively time and resource consuming (\cf  \Cref{sec:dataset-test}).
% ; not to mention the effort invested in training multiple annotators . 
Thus, we instead propose a weakly supervised method which learns from pseudo labeled data. We generate pseudo labels using existing NLP and vision techniques and then use these pseudo labels for training our model. We discuss this in detail below.

\subsection{Pseudo Label Generation}
\label{subsec:pseudo-label-gen}

\textbf{Event Detection in Text and Video.} The first step is detect events in text and video separately. We use the same automatic methods to detect them as used on the test data: Open Domain IE \cite{shen-etal-2021-corpus} and open source library PySceneDetect \footnote{\label{footnote:pyscene}\url{http://scenedetect.com/en/latest/}} for text event and video event detection respectively.

\textbf{Textual Hierarchical Relation Detection.} Assume we detected $m$ text events, $\{e_i\}_{i=1}^m$, in an article $T$ and $n$ video events, $\{v_j\}_{j=1}^n$, in the accompanying video $V$. The next step is to detect hierarchical relations among the text events, using \cite{wang-etal-2021-learning-constraints}, from all possible $m \times m$ pairs. Let $\{e_ue_{u_s}\}_{u=1,s=1}^{u=p,s=q}$ denote the hierarchically related event pairs, where the parent event is $e_u$ and the subevent is $e_{u_s}$ and $p,q \leq m$.

\textbf{Video Event Retrieval} The final step is to retrieve video events, $\{v^{u_s}_l\}_{l=1}^r$ and $r\leq n$, from video $V$, which depict the same real world event as the text subevent, $e_{u_s}$. This step essentially simplifies to a video retrieval task. As CLIP \cite{clip-2021} model has demonstrated state-of-the-art performance in multimodal retrieval tasks \cite{clip4clip-2021, fang2021clip2video}, we use it for this step. We provide more details in 
%Appendix. 
\Cref{appendix:pseudo-label-gen}.

We use CLIP to get all possible video events which are identical to the text subevent, denoted $\{e_{u_s}v^{u_s}_l\}$.
Critically, since $e_u$ was the parent event of $e_{u_s}$ and $e_{u_s}$ depicts the same event as $v^{u_s}_l$, we can conclude that $e_u$ is the parent event of $v^{u_s}_l$ by transitivity.
As a result, we get a total of 
% $\{e_uv^{u_s}_l\}_{u=1,s=1,l=1}^{u=p',s=q',l=r}$ 
$\{e_uv^{u_s}_l\}$
hierarchical event pairs and 
% $\{e_{u_s}v^{u_s}_l\}_{u=1,s=1,l=1}^{u=p',s=q',l=r}$ 
$\{e_{u_s}v^{u_s}_l\}$
identical event pairs. 

We collect additional identical pairs by directly comparing all text events $\{e_i\}_{i=1}^m$ in the article $T$ to all video events in $\{{v_j}\}_{j=1}^n$ in the paired video $V$ using CLIP. This gives an aggregate of 
% $\{e_{u_s}v^{u_s}_l\}_{u=1,s=1,l=1}^{u=p',s=q',l=r} \cup \{e_iv_j\}_{i=1,j=1}^{i=m',j=n'}$, where $m'<m, n'<n$, 
$\{e_{u_s}v^{u_s}_l\} \cup \{e_iv_j\}$
identical pairs.

In total, we collect 57,910 multimodal hierarchical event pairs and 39,0149 multimodal identical event pairs from the 100K video-article pairs training set. We evaluate the quality of these pseudo labels in \Cref{appendix:pseudo-label-quality}.

\begin{figure}[t]
    \centering
    \includegraphics[width=\linewidth]{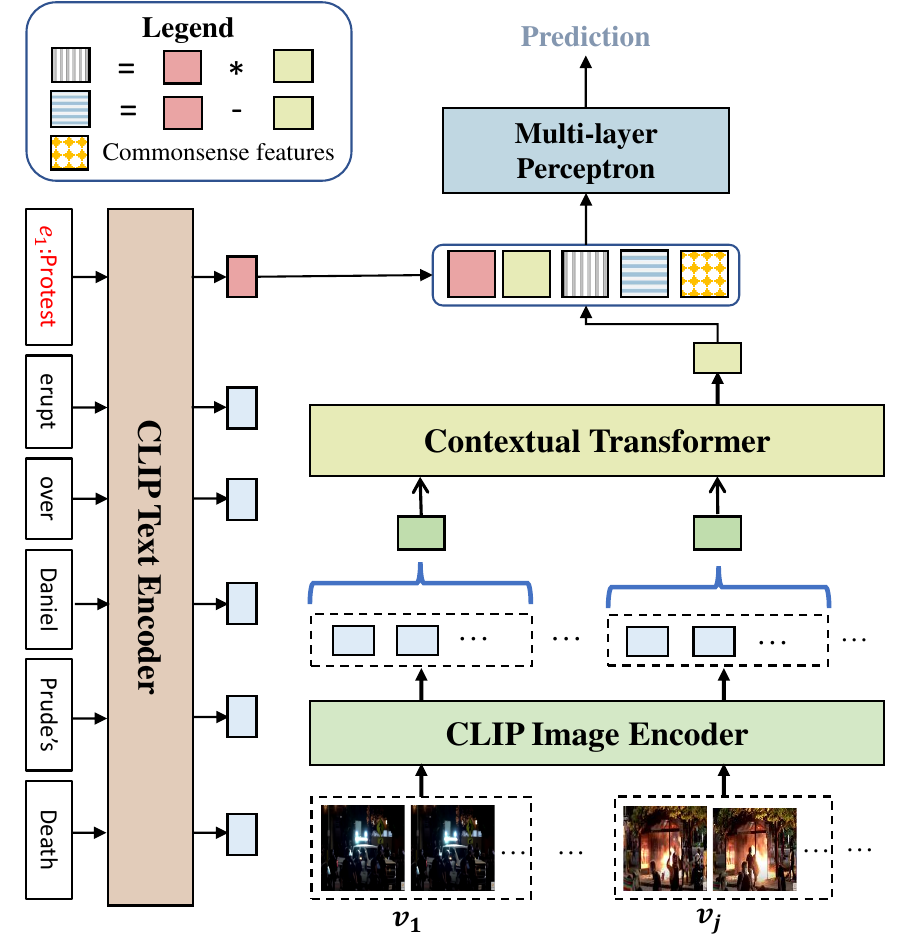}
    \caption{Overview of Proposed MASHER Model}
    % . $e_i$ denotes $i^{th}$ text event and $v_j$ denotes $j^{th}$ video event.}
    \label{fig:model}
\end{figure}

\subsection{Training}
\label{subsec:training}

Once we obtain the pseudo labels, we proceed to training using our model, \textbf{M}ultimodal \textbf{A}nalysis of \textbf{S}tructured \textbf{H}ierarchical \textbf{E}vents \textbf{R}elations (MASHER). The method is illustrated in \Cref{fig:model}. Given a text event $e_i$ and video event $v_j$ having a label from the pseudo label set, $r_{ij}'$, we follow the procedure described below to train our model.

\textbf{Input Representation and Feature Extraction.} We represent text events as a word, $e_i$, in a sentence $se_i=[w_1,w_2,...e_i,..w_j,...w_n]$. The video event, $v_j$, is a video clip in a video consisting of $n$ video events, $\{v_j\}_{j=1}^n$. $v_j$ is comprised of a stack of frames sampled uniformly at $f_s$ frames per second, $v_j=\{F_y^j\}_{y=1}^Z$. 
We use CLIP to extract text event features, $ft_i = \mathit{f'}_t(se_i)$ as well as video event features, $fv_j = \frac{1}{Z}\sum_{y=1}^Z \mathit{f}_i(F^j_y)$, where $f_i$ is CLIP's image encoder and $\mathit{f'}_t$ is a modification of CLIP's text encoder to capture additional context, $\mathit{f}_t$ (\cf  \Cref{appendix:pseudo-label-gen}).

\begin{figure*}[t]
    \centering
    \includegraphics[width=\linewidth]{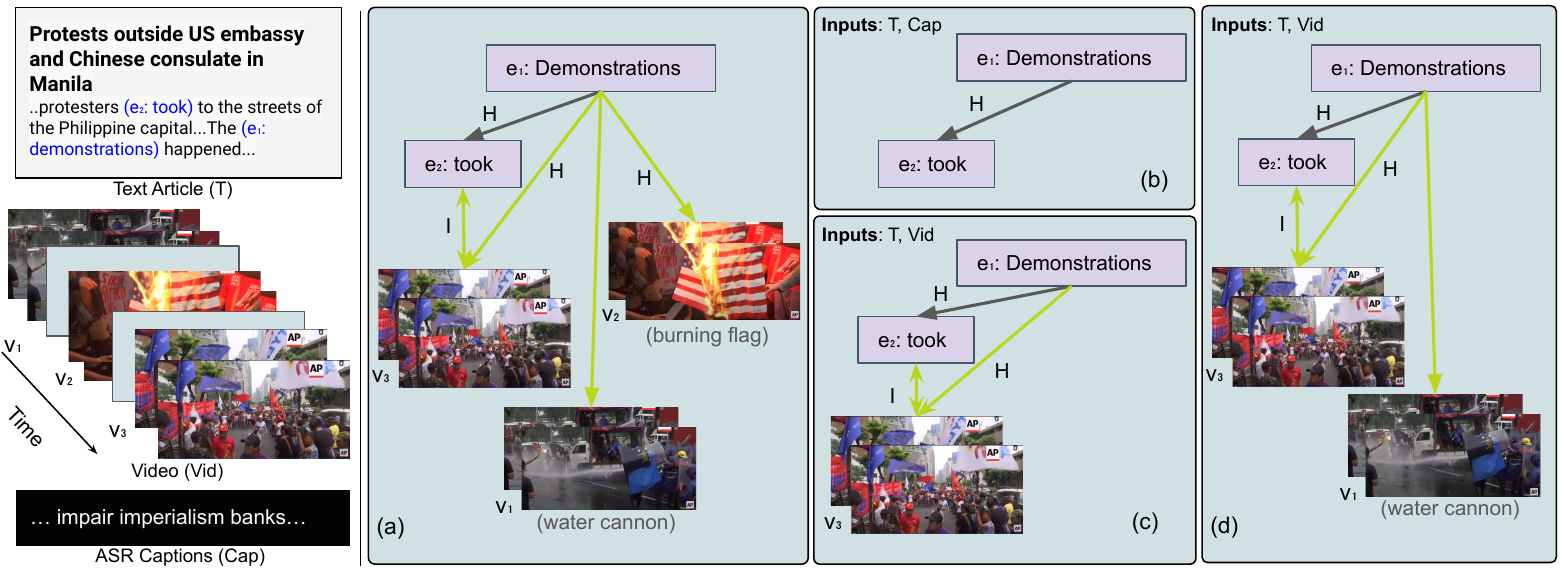}
    \captionof{figure}{The left most column shows the inputs and the rest are outputs. (a): Ground Truth, (b) Text Base. (c) MM Base. (d) MASHER. The text-text event relations are derived using the method described in \Cref{subsec:pseudo-label-gen}.}
    % \chris{The box for broken is cut off.}}
    \label{fig:qual-results}
\end{figure*}
% Please add the following required packages to your document preamble:
% \usepackage{multirow}
% \usepackage[table,xcdraw]{xcolor}
% If you use beamer only pass "xcolor=table" option, i.e. \documentclass[xcolor=table]{beamer}
\begin{table*}[t]
\centering
\small

\begin{tabular}{l ccc ccc c}
\toprule
                                        & \multicolumn{3}{c}{Hierarchical}                                       & \multicolumn{3}{c}{Identical}                                          &   \multirow{2}{*}{Avg $F_1$}                       \\
                     \cmidrule(lr){2-4} \cmidrule(lr){5-7}   
                                        & P & R & $F_1$ & P & R & $F_1$ &  \\ 
\toprule
Prior Base.                  & 4.7/2.0                  & 4.7/2.0                 & 4.7/2.0                    & 2.0/1.2                  & 2.0/1.2                 & 2.0/1.2                    & 3.4/1.6                                             \\
Text Base.               & 5.9/2.1                  & 0.1/0.1                  & 0.1/0.1                    & 2.5/2.6                  & 7.1/13.6                 & 3.6/4.3                    & 1.9/2.2                                             \\
MM Base. & \textbf{35.7}/\textbf{28.0}                 & 5.0/6.3                  & 8.8/10.3                    & \textbf{8.8}/\textbf{7.6}                  & 33.1/32.3                 & \textbf{13.9}/\textbf{12.4}                   & 11.4/11.4 \\

Video-LLaMA & 4.82/2.21 & 13.15/13.28 & 7.06/3.79  &  1.76/1.03 & 4.08/4.25 & 2.46/1.65 & 4.76/2.72 \\ \midrule

MASHER &   21.9/11.9 & \textbf{22.1}/\textbf{18.8} &	\textbf{22.0}/\textbf{14.6} &	8.2/6.3	& \textbf{44.5}/\textbf{39.0}	& \textbf{13.9}/10.9	& \textbf{18.0}/\textbf{12.8}  \\   
\bottomrule
\end{tabular}
\caption{Comparison with baseline models on the validation/test set.}
\label{tab:baselines-val}
\end{table*}

\textbf{Contextualizing Video Event Features} So far, we have extracted video event features independent of other events in the video. 
% But, this is limiting because a video event such as `building \textit{destruction}' could have happened because of a \textit{'storm'} event or an \textit{`earthquake'} event. \chris{****I DON'T UNDERSTAND THE PRIOR SENTENCE, OMIT?****}
This is a limitation since a video event such as building \textit{destruction} needs to be contextualized with respect to other events in the video to ascertain whether it happened because of, say, a \textit{``storm''} event or a \textit{``earthquake''} event.
As such, we use Contextual Transformer (CT) to contextualize the event features with respect to other events in the video.
CT is essentially a stack of multi-headed attention layers \cite{vaswani2017attention}. All the video events' features from video $V$, $\{fv_j\}_{j=1}^n$, forms the input tokens to CT. 
% We get the resulting contextualized features of video event $v_j$ as 
The output is $cfv_j=CT(\{fv_j\}_{j=1}^n)$.

\textbf{Commonsense Features}
To aid learning the relationship between open domain text and video events, we incorporate commonsense knowledge from an external knowledge base, ConceptNet \cite{conceptnet-2017}. Inspired by \cite{wang-etal-2020-joint}, we extract events related by relations ``HasSubevent'', ``HasFirstSubevent'' and ``HasLastSubevent'' from ConceptNet as positive pairs and random events as negative pairs. We embed the event pairs using CLIP and then leverage the embeddings to train a feature extractor $CS(.,.)$, a MLP (Multi Layer Perceptron), using contrastive loss 
%(\cf  Appendix).
(\cf  \Cref{appendix:method-train}). 
Once trained, we freeze it and use it as a commonsense feature extractor while training MASHER, $cs_{ij}=CS(ft_i, cfv_j)$. Although while training MASHER, one of the events is from the visual modality, we are still able to use $CS$ 
%as it is because we extract embeddings for the video event using CLIP as well and 
because CLIP's image embeddings and text embeddings lie in the same embedding space. We provide more analysis on this hypothesis in \Cref{appendix:method-train}.

\textbf{Embeddings Interactions (EI)} Following \cite{wang-etal-2020-joint}, we also add additional text event and video event feature interactions for a better representation. Specifically, (1) Subtraction of events' features (SF), $sf_{ij}=ft_i-cfv_j$ and (2) Hadamard product of events' features (MF), $mf_{ij}=ft_i * cfv_j$.

\textbf{Multi Layer Perceptron (MLP) \& Loss} We concatenate the text event feature, $ft_i$, contextualized video event features, $cfv_j$, commonsense features, $cs_{ij}$, and embedding interactions, $sf_{ij}$ and $mf_{ij}$, to form the input to a 2 layer MLP. The MLP is a 3-way classifier, outputting $p_{ij} \in \mathbb{R}^{1\times 3}$: the probabilities for $e_i$ and $v_j$ being classified as ``Hierarchical'', ``Identical'' or ``NoRel'' (Not Related). 
% The output is represented as $p_{ij} = \textrm{MLP}([ft_i; cfv_j; cs_{ij}; sf_{ij}; mf{ij}])$, where $p_{ij} \in \mathbb{R}^{1\times 3}$. 
We train the model using cross entropy loss between $p_{ij}$ and the label, $r_{ij}'$.

% \subsection{Inference}
% During inference, we use CLIP with MASHER as an ensemble to prune false positives for identical relations. 
% The rationale is to leverage CLIP's strong multimodal feature matching ability to prune out multimodal event pairs
% %some of the video and text events 
% falsely predicted as identical which CLIP can confidently invalidate. 

\subsection{Implementation Details}
 %(\Cref{appendix:impl-details}). 
We make a note that the majority of text event and video event pairs are not related (94.52\% in train set). To address this label bias, we weight the the labels in the cross entropy loss by the inverse ratio of their count in train set as done by \cite{wang-etal-2021-learning-constraints}. Our best model uses a single layer of multi-headed attention in CT. We train our model for 15 epochs using a batch size of 1024 and a learning rate of 1e-5 on 4 NVIDIA Tesla v100 GPUs for a total training time of around 34 hours. 
During inference, we use CLIP with MASHER as an ensemble to prune false positives for identical relations. 
The rationale is to leverage CLIP's strong multimodal feature matching ability to prune out event pairs
%some of the video and text events 
falsely predicted as identical, which CLIP can confidently invalidate.
We provide ablation study of our model architecture in \Cref{sec:results} and additional details on training hyperparameters in 
%Appendix.
\Cref{appendix:ablation-study}.

%-------------------------------------------------------------------------

\section{Experiments}

\textbf{Evaluation Metric}
% Following prior hierarchical event relations work in NLP \cite{hovy-etal-2013-events, glavavs2014hieve} and related scene graph \cite{xu2017scene} work in vision, we evaluate hierarchical and identical relations using Precision (P), Recall (R) and $F_1$ score (details in 
%Appendix).
% \Cref{appendix:eval-metric}).
% Recall/precision measures the percent of ground truth relations retrieved and prediction noise respectively. The resulting $F_1$ score adequately rewards the model for retrieving correct Hierarchical/Identical relations and penalizes them for making too many (wrong) predictions, making it a suitable metric for our task. We also report the macro average of $F_1$ scores of hierarchical and identical relations.
We evaluate hierarchical and identical relations using Precision, Recall, and F1-score, following prior work in NLP \cite{hovy-etal-2013-events, glavavs2014hieve} and related scene graph work in vision \cite{xu2017scene} (details in \Cref{appendix:eval-metric}). The F1-score adequately rewards the model for retrieving correct hierarchical/identical relations and penalizes them for making too many incorrect predictions. We also report the macro average of F1-scores of hierarchical and identical relations.

\subsection{Baselines}
\label{sec:baselines}

\textbf{Prior Baseline.} (Prior Base.)
We use a random weighted classifier which randomly predicts a relation type $\in$ \{``Hierarchical'', ``Identical'', ``NoRel''\} between any text and video event based on the prior distribution of the relation type in the annotated labels.

\textbf{Text-only Baseline.} (Text Base.)
We construct a text-only baseline to study the limitations of text-only data in this task.
% To evaluate the contribution of visual data in this task, we compare our model against a text-only baseline. 
To this end, we use ASR provided with video as a proxy for video events 
%(\cf  Appendix).
(\cf  \Cref{appendix:baselines}). 
Specifically, the proxy for video event $v_j$ is the ASR found within the timestamps of $v_j$, denoted $X_j$. 
We extract events from $X_j$, $\{e'_{j_z}\}_{z=1}^w$ following \Cref{subsec:pseudo-label-gen}. 
Next, we use the NLP model \cite{wang-etal-2021-learning-constraints} to predict the relationship type, $r_{ij_z}$ between a text event from the article, $e_i$, and proxy video events from ASR, $e'_{j_z}$. 
%found in video event $v_j$. 
If any $r_{ij_z}$$\in$ \{\textrm{``Hierarchical'', ``Identical''}\},
% \forall z$, 
we propagate the relation $r_{ij_z}$ from $e_i$ and $e'_{j_z}$  to $e_i$ and $v_j$ as $e'_{j_z}$ is a proxy for $v_j$.

\textbf{Multimodal Baseline.} (MM Base.)
We discussed a method to predict multimodal relations in \Cref{subsec:pseudo-label-gen}, which used NLP and vision methods to produce pseudo labels. This is currently the best performance that NLP and vision can separately combine to give without a trained model. 
As such, we consider this pipeline as our multimodal baseline.
% As such, we compare the performance of our model against this multimodal baseline, \chris{\ie ~the baseline used to produce its pseudo labels.} 
% We apply this baseline on both val and test sets to predict relations and evaluate it against the ground truth labels (Table~\ref{tab:baselines-val}).

% \textbf{BLIP-2} We input text event and the middle frame from video event to BLIP-2 \cite{li2023blip2} and prompt it with the definition of the task and the three types of relations.

\textbf{Video-LLaMA} \cite{zhang2023videollama}. It is a video based large language model which has demonstrated strong zero-shot results on multiple video language tasks. Consequently, we consider it as one of our baseline.

\subsection{Results}
\label{sec:results}

% \begin{figure*}[t]
%     \centering
%     \begin{subfigure}[t]{0.49\textwidth}
%     \centering
%     \includegraphics[width=\textwidth]{images/m2e2r_heatmap1.pdf}
%     \end{subfigure}
%     \rulesep
%     \begin{subfigure}[t]{0.49\textwidth}
%     \centering
%     \includegraphics[width=\textwidth]{images/m2e2r_heatmap2.pdf}
%     \end{subfigure}
%      \caption{Two model predictions (left and right): The left column shows the input, middle column shows the model output and the right column shows heatmap visualizations of the input based on the prediction. Legend: \textcolor{green}{Green} arrow = true positive, \textcolor{red}{Red} arrow = false negative.}
%     \label{fig:appendix-qual-samp}
% \end{figure*}

% Please add the following required packages to your document preamble:
% \usepackage{multirow}
% \usepackage[table,xcdraw]{xcolor}
% If you use beamer only pass "xcolor=table" option, i.e. \documentclass[xcolor=table]{beamer}
\begin{table}[t]
\centering
% \small
\footnotesize
\setlength{\tabcolsep}{4pt}

% ``SF'' denotes the substraction of textural and visual features. ``MF'' denotes the element-wise product of textural and visual features.}

\begin{tabular}{l rrr rrr r}
\toprule
 & \multicolumn{3}{c}{Hierarchical}                                       & \multicolumn{3}{c}{Identical}                                          &            \multirow{2}{*}{Avg $F_1$}   \\
 \cmidrule(lr){2-4} \cmidrule(lr){5-7}
 & \multicolumn{1}{c}{P} & \multicolumn{1}{c}{R} & \multicolumn{1}{c}{$F_1$} & \multicolumn{1}{c}{P} & \multicolumn{1}{c}{R} & \multicolumn{1}{c}{$F_1$} & \\
\toprule
MASHER Basic & 12.1                 & \textbf{30.7}        & 17.4                  & 7.2                  & 42.8                 & 12.4                  & 14.9                                            \\
\quad + CT & 17.4                 & 23.9                 & 20.1                  & 7.5                  & 43.3                 & 12.8                  & 16.5                                            \\
\quad + CS & 13.4                 & 29.1                 & 18.4                  & 7.6                  & 44.9                 & 13.0                  & 15.7                                            \\

\quad + EI & 15.2                 & 29.3                & 20.0                  & 7.3             & \textbf{45.1}                 & 12.5                  & 16.3                                            \\

\quad + CT + CS + EI & \textbf{21.9}        & 22.1                 & \textbf{22.0}         & \textbf{8.2}         & 44.5                 & \textbf{13.9}         & \textbf{18.0} \\                       
\bottomrule
\end{tabular}
\caption{Ablation studies on components and features.}
% ``MASHER Basic'' is the basic MASHER model. ``CT'' denotes the contextual transformer. ``CS'' denotes the common-sense knowledge features. ``EI'' denotes the embedding interactions, including the subtraction and element-wise product of textural and visual features.}
\label{tab:appendix-network-arch}
\end{table}

\textbf{Comparison against baselines}
\label{sec:baseline-comp}
The comparison between MASHER and above-mentioned baselines on the validation and test set are reported in Table~\ref{tab:baselines-val}. We also compare MASHER's and the baselines' performance on a dataset sample visually in \Cref{fig:qual-results}. From the table and the figure, we make following observations: 

\begin{itemize}[leftmargin=*]
    % \item MM Base. performs better than Text Base. demonstrating the importance of \chris{visual data to the task of predicting cross-modal event relations.}
    
    \item For the most comprehensive metric, Avg $F_1$ score, MASHER outperforms all baselines with significant performance gains (\eg  18.0 vs. 11.4 on the validation set).
    
    \item Text Base. performs quite poorly (Avg $F_1$ 2.2). This is because a lot of visual events in the video are not mentioned in its ASR. This fact is also demonstrated by \Cref{fig:qual-results}.
    
    \item MM Base. performs better than Text Base. (Avg $F_1$ 11.4 vs 2.2), stressing the importance of visual data to this task.
    
    \item MASHER achieves 4x recall over MM Base for hierarchical relations. 
    % It is because MASHER can additionally predict visual-only subevents as well.
    It is because MM Base. relies on finding the subevent in text first before retrieving the matching video subevent (\Cref{subsec:pseudo-label-gen}). This causes it to miss visual-only subevents while MASHER can discover those as it directly predicts multimodal relations.
    This fact is evident in \Cref{fig:qual-results} -- MM Base can only discover ``took'' (to streets) video subevent as it is mentioned in text as well. While MASHER can also detect ``water cannon'' visual-only subevent.
    % directly predict multimodal hierarchical relations from text parent events to video subevents.
    % In contrast, MM Base. relies on finding the subevent in text first before retrieving the matching video subevent (\Cref{subsec:pseudo-label-gen}). 
    % Absence of the subevent mention in text causes it to miss numerous relations. 
    We further validate this hypothesis by measuring recall on visual-only subevents. MASHER scores 15.92\% while MM Base. scores 2.14\%.
    This also explains MM Base's better precision, since it only predicts a few relations.
    
    % \item Precision for identical relations is low for both MM Base. and MASHER. This is because they both use a video retrieval component in their pipeline which marks hierarchical relations as identical. For example, a ``meeting'' text event is marked identical to a video subevent showing a handshake. This is due to the nature of the video retrieval model's training dataset. Whereas, in our dataset, handshake is only one aspect of a ``meeting'' and thus is annotated hierarchically.
    \item Both MM Base and MASHER have low precision for identical relations due to their use of a video retrieval component that noisily predicts hierarchical relations as identical. For instance, a "meeting" text event is predicted identical to a video sub-event showing a handshake, because of the nature of the training dataset used for the video retrieval model. 
    % In contrast, our dataset annotates handshake as only one aspect of a "meeting," which results in hierarchical annotation.
    In contrast, our dataset annotates handshake to be a subevent of ``meeting'', as it's only a part-of meeting event.
    \item Video-LLaMA, being a strong multimodal baseline, outperforms both Prior and Text Base. However, it's worse than MM Base, indicating specialized models (MM Base.) perform better than generic vision-language model (Video-LLaMA) on this task.
\end{itemize}
% 1) For the most comprehensive metric Avg $F_1$ score, MASHER obviously outperforms all baselines with significant performance gains (\eg  18.0 vs. 11.4 on the validation set). 
%2) For Hierarchical relation type, our model achieves 4x recall over multimodal baseline (MM Base.). 
%much higher recall scores (\eg  22.0 vs. 8.8 and 14.6 vs. 10.3 for hierarchical relations). 
% This is because our model can directly predict multi-modal `Hierarchical' relationship from text parent event to video subevent. In comparison, MM Base. relies on finding the subevent in text first before retrieving the matching video subevent (\Cref{subsec:pseudo-label-gen}). Absence of subevent mention in text causes it to miss numerous multimodal event relations. Meanwhile, since the multimodal baseline only produces quite a few predictions, it achieves higher precision scores. 

% We also note the performance improvement of MM Base. over Text Base. demonstrating the criticality of visual data to this task.

% \begin{figure}[t]
%     \centering
%     \includegraphics[width=\linewidth]{images/heatmap2_2.pdf}
%      \caption{GRAD-CAM heatmap visualizations of the input based on MASHER's prediction. \textcolor{OliveGreen}{$\rightarrow$}: true positive, \textcolor{red}{$\rightarrow$}: false negative.}
%     \label{fig:heatmap}
% \end{figure}

\textbf{Ablation Study and Analysis}
%\textbf{Effectiveness of Different Features} 
% We conduct a set of ablations to verify the importance of different features used in our model and report the results
%including the contextual transformer feature (CT), commonsense feature (CS), and embedding interactions (ET). 
% in Table~\ref{tab:appendix-network-arch}.
In Table~\ref{tab:appendix-network-arch}, we report the results of ablation study done to verify the importance of different features used in our model.
We make several observations from this: 1) contextualizing video event features with respect to other video events (using CT) helps; 2) an external Knowledge base (via CS features) improves understanding of open domain event-event relations; 3) different embedding interaction techniques (EI) helps improve feature representation; and 4) all three network components (CT, CS and EI) synergistically combine to give the best performance. 
% We further ablate the effect of inference time ensemble with CLIP in \Cref{tab:identical-pruning}. As expected, it improves performance by eliminating identical false positives. 
We study more ablations on the effect of inference time ensemble with CLIP and the number of layers in CT in \Cref{appendix:ablation-study}.
% Further, through Grad-CAM \cite{gradcam, jacobgilpytorchcam} heatmap visualizations in \Cref{fig:heatmap}, we illustrate that MASHER pays attention to relevant portions of video frames while predicting event-event relationships. More illustrations in \Cref{appendix:qual-analysis}.
Further, we show that MASHER pays attention to relevant objects through Grad-CAM \cite{gradcam} heatmap visualizations in \Cref{appendix:qual-analysis}.

\section{Conclusion}

We proposed the novel task of extracting multimodal event hierarchies from multimedia content, a powerful way to understand, represent and reason about our world. Along with the task, we introduced {\fontfamily{qcr}\selectfont MultiHiEve} -- a video-language dataset sourced from news domain and containing rich hierarchy of events. 
% We annotated a portion of this dataset for benchmarking and evaluation. 
We proposed a weakly supervised model, MASHER, to predict these multimodal event relationships, achieving an improvement of around 3x on recall and 50\% on $F_1$ score (hierarchical relations) over the strongest baseline.
% without requiring any annotations for training. While our method makes significant advances on the proposed task, the performance of 12.8 Avg $F_1$ score underlines the immense difficulty of the task and consequently provides fertile ground for future work.

% Our work has several limitations which makes for exciting explorations in the future -- 1) discovering hierarchical relations from video to text; 2) exploring temporal and causal relations; and 3) leveraging more noise-aware learning strategies to deal with noisy pseudo labels. 
% 2) Additionally, relying on grounding methods in our pseudolabeling procedure results in some hierarchical relations being mislabeled as identical. 
% Future work may address these by further exploiting the transitivity of text-video relations for supervision and leveraging more robust noise-aware learning strategies. %We analyse the limitations and suggest future directions of our work in \Cref{appendix:limitations}. 

We discuss the limitations and future directions of our work in \Cref{appendix:limitations}. We also discuss privacy and social bias concerns with respect to {\fontfamily{qcr}\selectfont MultiHiEve} 
% and the steps we have taken to mitigate them 
in \Cref{appendix:dataset-licensing}.
% As large scale datasets tend to have privacy, distribution and social bias concerns, we discuss these issues with respect to {\fontfamily{qcr}\selectfont MultiHiEve} and the steps we have taken to mitigate them in \Cref{appendix:dataset-licensing}. We also include the Dataset sheet \cite{Gebru2021DatasheetsFD} describing additional details of our dataset in \Cref{appendix:dataset-sheet}.

% As is the case with any large scale dataset, there are privacy, distribution and social bias concerns with {\fontfamily{qcr}\selectfont MultiHiEve}. We discuss these concerns below and the steps we have taken to minimize them.

% Downloading YouTube videos risks compromising privacy of individuals. To address this, we have only downloaded videos from the official YouTube channel of public news entities. To mitigate licensing issues, we only release urls of the videos following \cite{merlot-2021, miech2019howto100m}. Further, we believe academic usage of the data constitutes `fair use'.

% As the domain of our dataset is news, it risks being opinionated and, at times, biased towards certain politcial ideology, race, gender, culture, society and religion. We have actively taken steps to minimize such biases by selecting `Center' rated media sources (\cref{sec:dataset}). We provide more details in \Cref{appendix:dataset-licensing}.

% {\small
% % \bibliographystyle{ieee_fullname}
% \bibliographystyle{abbrvnat}
% \bibliography{egbib}
% }

\section*{Acknowledgement}

This research is based upon work supported by U.S. DARPA
KAIROS Program No. FA8750-19-2-1004. The views and
conclusions contained herein are those of the authors and
should not be interpreted as necessarily representing the official policies, either expressed or implied, of DARPA, or
the U.S. Government. The U.S. Government is authorized to
reproduce and distribute reprints for governmental purposes
notwithstanding any copyright annotation therein.

\bibliography{aaai24}

\clearpage

\appendix

\section*{Appendix}

We provide additional details in this section which further elucidate the claims and contributions of the paper. It's divided into the following sections:

\begin{itemize}
    \item Further details on Events' definition (\Cref{appendix:event-def})
    \item Details on dataset curation, statistics, annotation and broader impacts. (\Cref{appendix:dataset})
    \item Details on various model components (\Cref{appendix:method})
    \item Details on evaluation metric and baselines (\Cref{appendix:exp-details})
    \item Ablation and Hyperparameter Tuning (\Cref{appendix:ablation-study})
    \item Qualitative Analysis on predicted results (\Cref{appendix:qual-analysis})
    \item System Evaluation including text event detection (\Cref{appendix:iete2ve})
    \item Limitations and Future Directions (\Cref{appendix:limitations})
    \item {\fontfamily{qcr}\selectfont MultiHiEve} Dataset sheet \cite{Gebru2021DatasheetsFD} (\Cref{appendix:dataset-sheet})
\end{itemize}

\section{Events and Events-Events Relations}
\label{appendix:event-def}

Multimodal event graphs consists of events (text and video) as nodes and relations between them as edges. For the purpose of clarity and consistency, we provide additional details on the definition of each type of events and their relations.

\subsection{Text Events}
\label{appendix:text_event_def}

Events in text denote a change of state or the changed state of a person, thing or entity. An event trigger is the word which most clearly expresses the occurrence of the event. For the purpose of annotating events in text article, we task the annotators to annotate all possible event triggers. Events triggers can be (underlined text denote event triggers):

\begin{itemize}
    \item Verbs. (eg. They have been \underline{married} for three years)
    \item Nouns. (eg. The \underline{attack} killed 7 and injured 20)
    \item Nominals. Nominals are created from verb or adjective, eg. decide $\rightarrow$ decision.
    \item To resolve the cases where both a verb and a noun are possible event triggers, we select the noun whenever the noun can be used by itself as a trigger word (eg. The leaders held a \underline{meeting} in Beijing).
    \item To resolve the cases where both verb and an adjective are possible event triggers, the adjective is selected as the trigger word whenever it can stand alone to express the resulting state brought about by the event (eg. The explosion left at least 30 \underline{dead} and dozens \underline{injured}).
    \item In cases where multiple verbs are used together to express an event, the main verb is chosen as the event trigger (eg. John tried to \underline{kill} Mary).
    \item In case the verb and particle occur contiguously in a sentence, we will annotate both the verb and particle as event trigger (Jane was \underline{laid} \underline{off} by XYZ Corp).
    \item Sometimes, multiple events are present withing a single scope ie. the sentence. In such cases, each sentence should be annotated with trigger words corresponding to all the events (eg. The \underline{explosion} left at least 30 \underline{dead}).
\end{itemize}

\begin{figure}
    \centering
    \includegraphics[width=\linewidth]{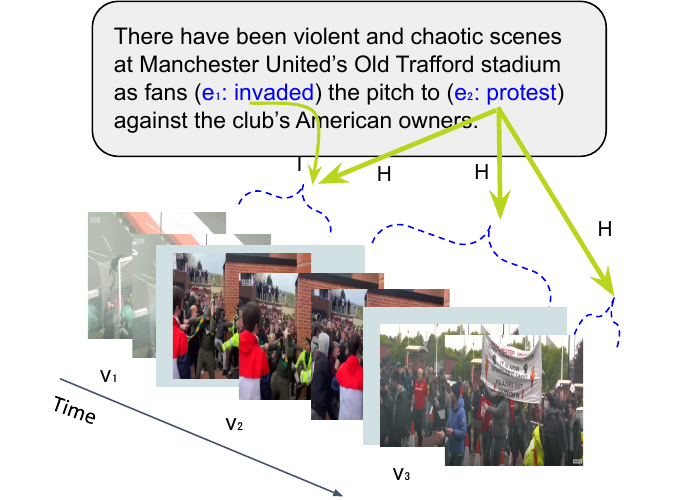}
    \caption{Illustration of Hierarchical and Identical Relation Types. I: Identical; H:Hierarchical. While event $e_1$: invaded is `Identical' to the video event showing people on the pitch $v_1$, event $e_2$: protest is `Hierarchical' to all video events -- $v_1$: showing invasion, $v_2$: showing clash, $v_3$: showing rally -- representing different aspects of protest.}
    \label{fig:rels}
\end{figure}

Additionally, we ask the annotators not to annotate the following type of events (underlined text denote events not to be annotated):

\begin{itemize}
    \item Negative Events. An event is negative when it is explicitly indicated that the event did not occur (eg. His wife was sitting on the backseat and was not \underline{hurt}). 
    \item Hypothetical Events. This includes the events which are believed to have happened or are hypothetical in nature (eg. Chapman would be concerned for his safety if \underline{released}).
    \item Generic Events. A generic event is any event which is contained in a generic statement (Terrorists often \underline{flee} to nation-states with crumbling governments to avoid \underline{interference}).
\end{itemize}

% \subsection{Shot segments as Video Events}
% % \todo{Show example of video shots as justification for video events}

% Generic event detection in videos is a hard problem \cite{genericEventBoundary2021}. It's hard to define what constitutes a video event both temporally and semantically. This leads to inconsistencies in annotation. 

\subsection{Event Relation Types}
% \todo{Show diagram showing different relation types}

\begin{figure*}
    \centering
    \vspace{-2.5em}
         \includegraphics[width=\textwidth]{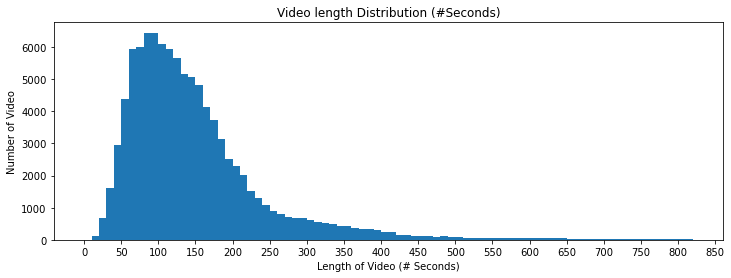}
         \includegraphics[width=\textwidth]{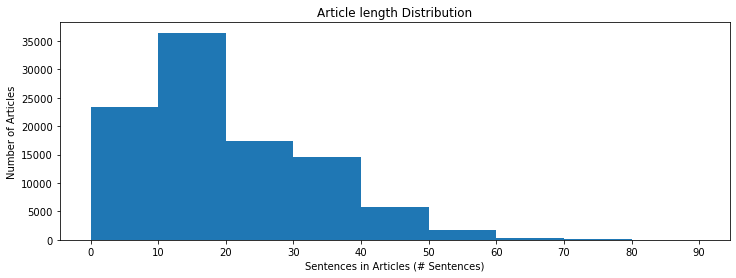}
         \includegraphics[width=\textwidth]{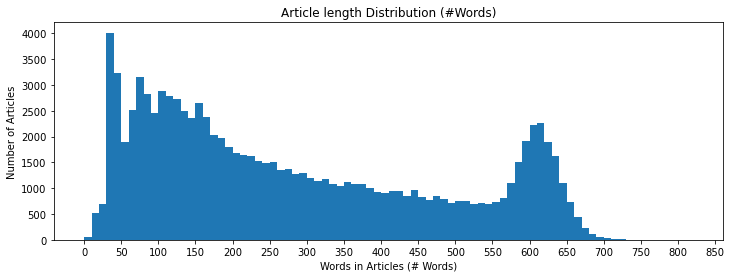}
     \caption{Data Statistics: Top row shows the distribution of video duration, middle row shows the distribution of article sentence length and the last row shows the distribution of article Word length.}
    \label{fig:data-distr}
\end{figure*}

\begin{table*}[t]
\centering
\small

\begin{tabular}{l l rrrc cl}
\toprule
Dataset       & Domain  & \#Videos &  \#Total hours & ASR & \#Text Sent. & Text Type \\
\midrule
MSVD~\cite{chen2011collecting}         & Open    & 1970                                                                 & 5.3                                & \xmark  & 70K & Caption\\
MPII Cooking~\cite{rohrbach2012database}  & Cooking & 44                                                                  & 8                                  & \xmark  & 5609   & Caption \\
TACoS~\cite{regneri2013grounding}         & Cooking & 127                                                                 & 15.9                               & \xmark  & 18K & Caption\\
TACos-MLevel~\cite{rohrbach2014coherent}  & Cooking & 185                                                                 & 27.1                               & \xmark  & 53K   & Caption  \\
M-VAD~\cite{torabi2015using}         & Movie   & 92                                                                  & 84.6                               & \xmark  & 56K    & Caption  \\
MSR-VTT~\cite{xu2016msr}       & Open    & 7180                                                                 & 41.2                               & \xmark  & 200K      & Caption     \\
Charades~\cite{sigurdsson2016hollywood}      & Daily activities   & 9848                                                                 & 82.01                              & \xmark  & 16K   & Caption   \\
VTW~\cite{zeng2016generation}           & Open    & 18.1K                                                                & 213.2                              & \xmark  & 45K    & Caption        \\
YouCook II~\cite{ZhXuCoAAAI18}    & Cooking & 2000                                                                & 176                                & \xmark  & 15K     & Caption    \\
ActyNet Cap~\cite{krishna2017dense}   & Open    & 20K                                                                 & 849                                & \xmark  & 100K   & Caption     \\
\color{gray} HowTo100M~\cite{miech2019howto100m}& \color{gray} Instructional    & \color{gray} 1.2M                           & 
 \color{gray} 134K & \color{gray} \cmark & \color{gray} - & \color{gray} - \\   
\color{gray} YT-Temporal-180M\cite{merlot-2021}& \color{gray} Open    & \color{gray} 6M                           & \color{gray} - & \color{gray}  \cmark & \color{gray} - &\color{gray} - \\  
\midrule
{\fontfamily{qcr}\selectfont MultiHiEve} (Train) & News    & 100K                                                           & 4.1K                            & \cmark & 1.9M  & News Story  \\
\bottomrule
\end{tabular}
\caption{Comparison of {\fontfamily{qcr}\selectfont MultiHiEve} training data against related video-language datasets. Avg. len: Average length of videos in seconds. Text Sent: Text sentences. Datasets not containing text descriptions are greyed.}
\label{tab:appendix-train-dataset}
\end{table*}

There are primarily four types of relations that have been explored in the literature: identical \cite{hovy-etal-2013-events, glavavs2014hieve}, hierarchical \cite{wang-etal-2021-learning-constraints}, temporal \cite{wang-etal-2020-joint} and causal \cite{yao-etal-2020-weakly, Sadhu_2021_CVPR}. While each of these relations are important towards building a comprehensive event graph, exploring all of them together is extremely challenging due to the annotation time required and the difficulty at distinguishing the different relation types for annotators. Despite the event relation prediction being a longstanding and established problem in NLP, no single large scale dataset exists containing all four relation types. Because the annotation and task difficulty increases substantially in the multimodal setting, we focus on the two most frequently found multimodal relation types in our human annotated data: Hierarchical and Identical relations. 

Identical events are those events which denote exactly the same real world event. In contrast, two events are said to be hierarchically related if one of the event (parent event) spatio-temporally contains the other event (subevent). For this task, we constrain parent event to be a text event only to simplify the task and limit disagreements between the annotators. We provide additional illustration of the difference in these two types of relations in \Cref{fig:rels}.

As a corollary, a video event is `Identical' to a text event if and only if the text event begins and ends in the video event and there is no more aspect to it, otherwise the text event is `Hierarchical' to the video event. For example, a text event, \textit{`meeting'}, will have a `Hierarchical' relation to a video event showing two people shaking hands because the event \textit{`meeting'} has aspects other than just shaking hands like \textit{`discussion'}, \textit{`interview'} etc.

Although, the direction of our Hierarchical relation if from text event to video event, we make a note that there could be event-event relations directed from video event to text event. However, we do not deal with them in this work to make the task more tractable.

\section{Dataset Details}
\label{appendix:dataset}

\subsection{Dataset Curation}
\label{appendix:dataset-curation}

We consider the news domain for our dataset collection as news stories often contain rich events information along with their nuanced event-event relations. However, news articles are often opinionated and run the risk of being biased. One way of adjudging this bias is to rate the political leaning of news outlet as being left, center or right. To minimize such bias and collect only factual news, we employ media rating website, \texttt{allsides.com}, to select news agencies rated as `Center'. Using this strategy, we end up with 9 news media sources: Associated Press, Axios, BBC News, Christian Science Monitor, Forbes, NPR, The Hill, Wall Street Journal and Reuters.

Having finalized the news sources, we collect YouTube channel ids for all these sources. Next, we scrape videos, associated descriptions and closed captions from the collected channel ids. Videos longer than 14 minutes or having a description length smaller than 10 words were filtered out. We end up with dataset of size 100.5K samples: 100K for training and 526 for evaluation. We further divided this evaluation set into two sets: 249 for validation and 277 for testing.

\begin{figure*}[!ht]
\centering
   \begin{minipage}[t]{\textwidth}
        \centering

\small
\begin{tabular}{ll}
\toprule
White House     &       trump obama donald white house barack biden presidential republican  \\
International   &       talks peace agreement leaders secretary deal conference  \\
Attack          &       injured hospital car attack bomb blast wounded scene damaged  \\
Election        &       election vote elections presidential voters voting candidate percent \\
Justice         &       court trial judge case charges lawyer prison accused lawyers \\
War             &       soldiers troops army forces fighting border rebels area fire \\
Entertainment   &       film actor star movie actress director hollywood festival\\
Conference      &       conference cutaway reporters general spokesman want need  \\
Congress        &       house senate congress committee trump senator republican democrats  \\
Protest         &       protesters protest demonstrators anti chanting rally march   demonstration \\

\bottomrule
\end{tabular}
\captionof{table}{Top 10 topics derived from {\fontfamily{qcr}\selectfont MultiHiEve} represented by their most common words.}
\label{tab:dataset-topic-model}
   \end{minipage} 
  \begin{minipage}[]{0.45\textwidth}
    \includegraphics[width=\textwidth]{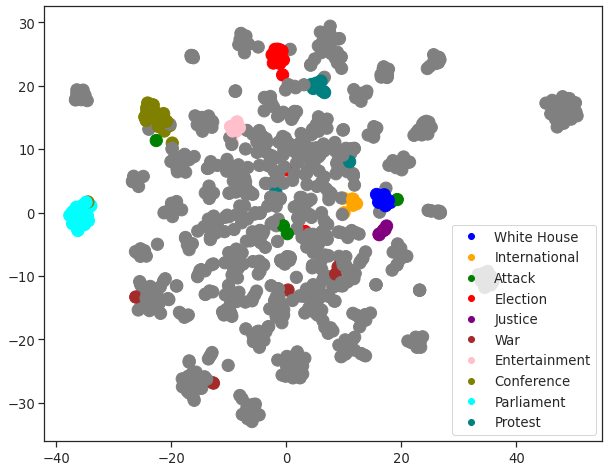}
    \caption{ \small{TSNE topic distribution for 1K random articles}}
      \label{fig:data-tsne}
  \end{minipage}
  \begin{minipage}[]{0.45\textwidth}
    % \vspace{-2.5em}
    \includegraphics[width=\textwidth]{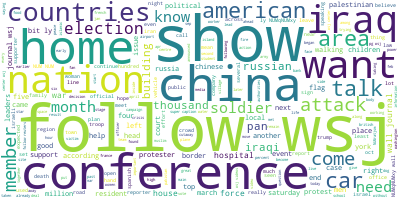}
    \caption{\small{Wordle of 1K random articles}}
      \label{fig:data-wordle}
  \end{minipage}
\end{figure*}

\subsection{Dataset Exploration}
\label{appendix:dataset-exploration}

\paragraph{Data Statistics} We provide additional data statistics on distribution of video duration, text article \#words and \#sentences in \Cref{fig:data-distr}. We observe that most of the videos in our dataset are at least 100 seconds long and most of the text articles have at least 10 sentences and 50 words. We compare these statistics against other video language dataset below.

We compare our dataset against 12 related video language datasets in \Cref{tab:appendix-train-dataset}. As can be seen from the table, our dataset is the first to be sourced from news domain which contains diverse event mentions and rich event hierarchies in the text article. From among the datasets containing text descriptions, the scale of our dataset is around 10x and 5x bigger in terms of the number of sentences and total video duration respectively. Datasets such as HowTo100M \cite{miech2019howto100m} and YT-Temporal-180M \cite{merlot-2021} are orders of magnitude larger than our dataset but they only contain closed captions as textual data without accompanying text descriptions/articles. Also, we are the only dataset that provides both text descriptions and ASR.

\paragraph{Data Topic Exploration}

To discover the range of news topics covered by our dataset, we employed LDA topic modeling \cite{Blei03lda} implemented in Mallet \cite{mallet} following \cite{merlot-2021}. Precisely, we grouped our dataset into 100 topics using a vocabulary of size 22K consisting of words which occurred in at least 25 articles but no more than 10\% of articles. Further, we removed those topic which contained media source names (eg. Reuters, WSJ etc.).

We report the top 10 topics in \Cref{tab:dataset-topic-model}. We see that a diverse range of topics are covered by our dataset: Election, Justice, War, International News, Entertainment etc. For a random 1000 data samples, we visualize the topic distribution via tsne plot in \Cref{fig:data-tsne} and a wordle of most prominent words in \Cref{fig:data-wordle}.

\subsection{Annotation Tool}
\label{appendix:dataset-annotation}

% \begin{figure}[t]
%     \centering
%     \begin{subfigure}[]{\textwidth}
%          \centering
%          \includegraphics[width=\textwidth]{images/m2e2r_annotations1.pdf}
%      \end{subfigure}
%      \begin{subfigure}[]{\textwidth}
%          \centering
%          \includegraphics[width=\textwidth]{images/m2e2r_annotation2.pdf}
%      \end{subfigure}
%      \begin{subfigure}[]{\textwidth}
%          \centering
%          \includegraphics[width=\textwidth]{images/m2e2r_annotations3.pdf}
%      \end{subfigure}
%      \caption{Wordle of 1K random articles}
%      \label{fig:annotation-tool}
% \end{figure}

% \begin{figure}
%     \centering
%     \includegraphics[width=0.9\textwidth]{images/m2e2r_annotations1.pdf}
%     \includegraphics[width=0.9\textwidth]{images/m2e2r_annotation2.pdf}
%     \includegraphics[width=0.9\textwidth]{images/m2e2r_annotations3.pdf}
%     \caption{Caption}
%     \label{fig:my_label}
% \end{figure}

% \begin{figure}
%     \centering
%     \includegraphics[width=\textwidth]{images/annotation.pdf}
%     \caption{Caption}
%     \label{fig:my_label}
% \end{figure}

For annotating the evaluation set, we used the Label Studio\footnote{\url{https://labelstud.io/}} annotation tool. We did not use the vanilla tool; rather we customized it according to our use case. The screenshots of annotation interface is shown in \Cref{fig:annotation-tool}.

The top row shows an overview of our interface. The text article, pre-populated with automatically extracted text events (depicted by red boxes), is on the left. The video along with a scroll bar and play/pause button is in the center. It has also been pre-populated with automatically extracted video events. These video events are shown by segmented green bars in center middle. The dark green segment shows the selected video event. The rightmost panel shows information on selected event, the list of all annotated text and video events and finally, the list of all event-event relation annotations.

The annotation task is divided into two subtasks -- 1) adding/deleting text events; and 2) annotating all possible `Hierarchical' and `Identical' multimodal relations from text event to video event. The middle rows shows the steps involved in adding a text event in blue and deleting a text event in pink. To add text event, the annotator first needs to click on the box marked `Text Event' at top left corner. Then, they needs to select word(s) from article to be labelled as a text event. To delete a text event, the annotator needs to first click on the red box containing the event to be deleted. Then, they need to either hit backspace key or click on the delete icon on the rightmost panel.

The bottom row illustrates the process of labelling a multimodal relation from text event to video event. To annotate a relation, the annotator needs to first click on the text event that they want to label the relation from. Then, they need to click on the relation icon on the rightmost panel. Next, the video event segment, to which the relation is directed to, needs to be clicked. Finally, the type is relation needs to be selected from the drop down menu displayed at the bottom on rightmost panel.

Once the annotation is complete, we export the labels in json format.

\begin{table*}[htbp!]
    \centering
    \scalebox{0.8}{
    \begin{tabular}{c}
    
    \includegraphics[width=\linewidth]{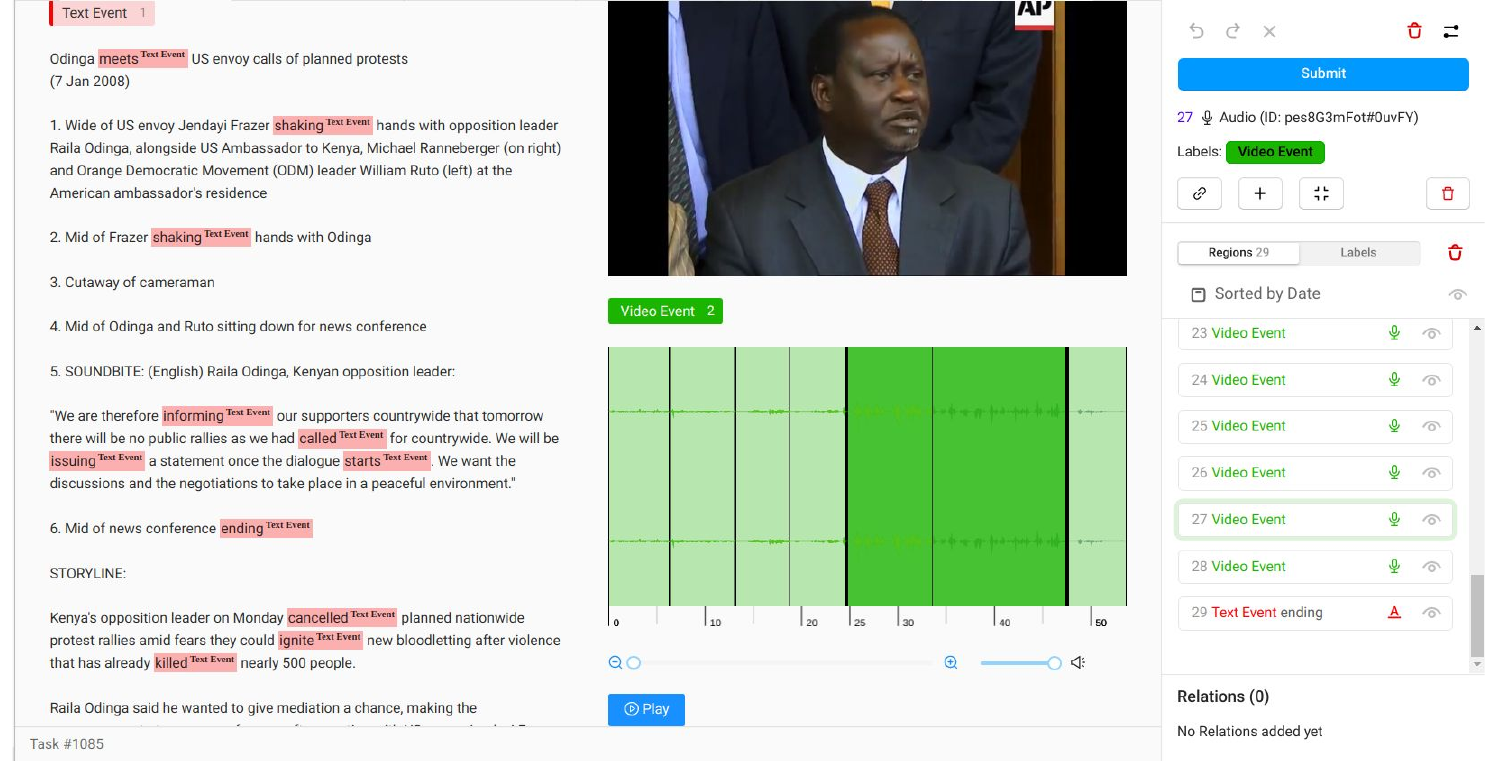} \\
    \bottomrule
    \includegraphics[width=\linewidth]{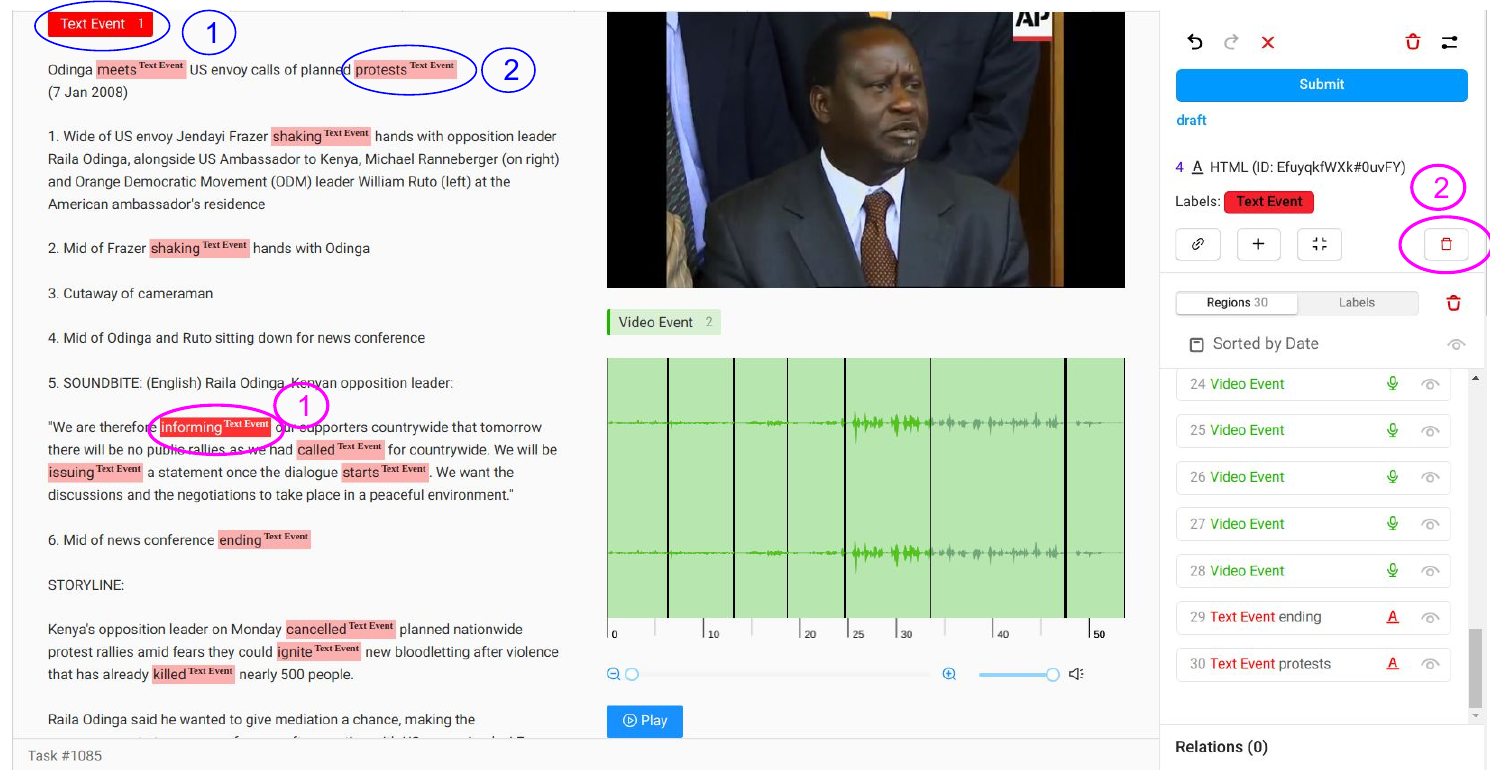} \\
    \bottomrule
    \includegraphics[width=\linewidth]{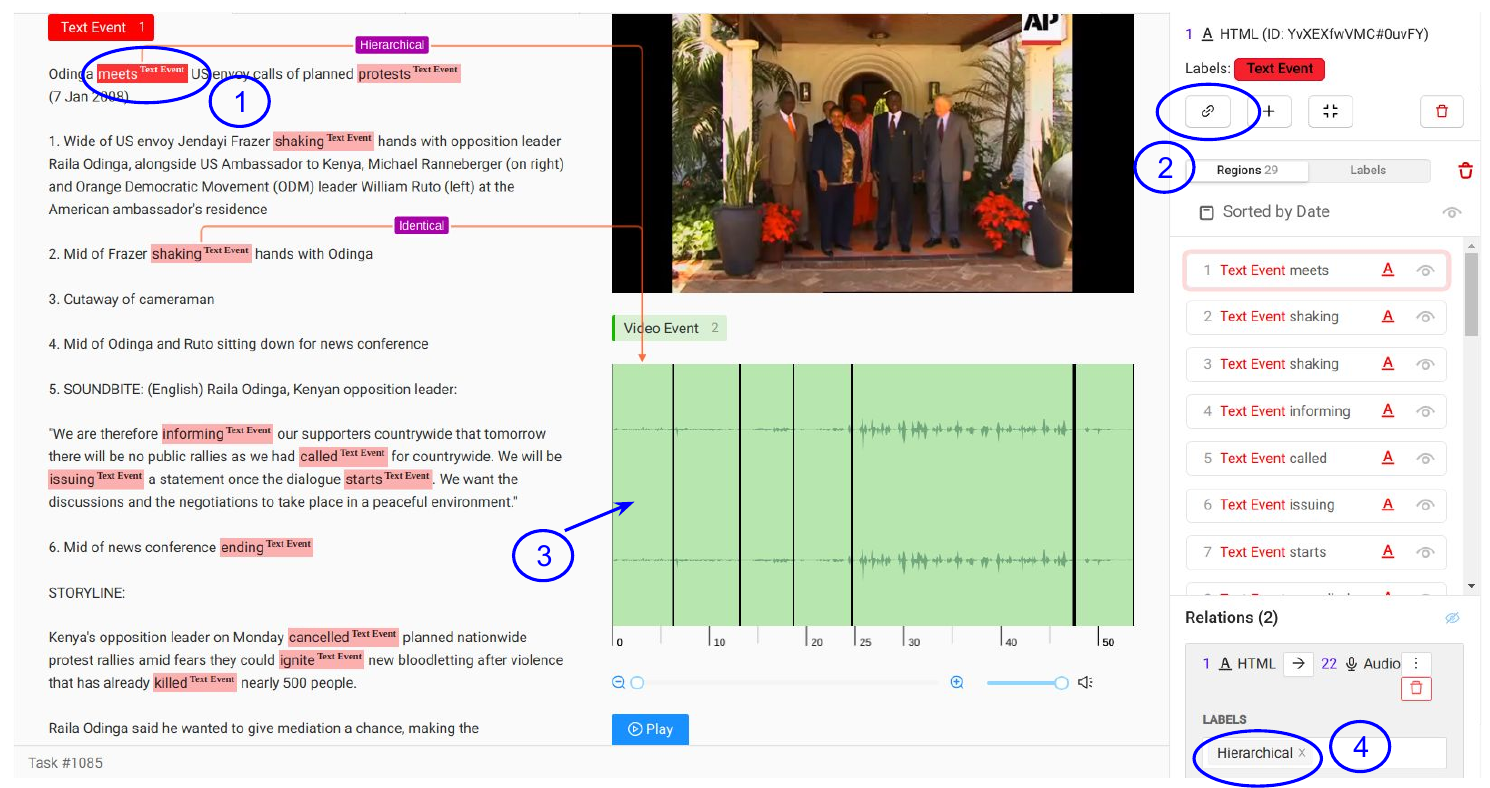} \\
    
    \end{tabular}
    }
     \captionof{figure}{Top: Overview of Annotation Tool. Middle: Task 1 -- Text Event Annotation. Steps for adding text event shown in {\color{blue}blue} and steps for deleting text event shown in {\color{magenta}pink}. Bottom: Task 2 -- Multimodal Event-Event Relation Annotation}
     \label{fig:annotation-tool}
\end{table*}

\subsection{Privacy, Distribution and Social Impact}
\label{appendix:dataset-licensing}

\paragraph{Distribution and Licensing}
Following data release strategy from prior work \cite{merlot-2021, miech2019howto100m}, we will be only releasing urls of the YouTube videos. This enables us to publicly release the data quickly while also minimizing potential consent and privacy issues. The usage of the data is for non-commercial research and educational purposes, which we believe constitutes 'fair use'. However, our data release strategy is aimed at further reducing any potential licensing issues.

\paragraph{Privacy} Downloading YouTube videos risks compromising privacy of individuals. To address this, we downloaded videos from the official YouTube channels of news media companies. As such, the videos fall under the category of journalism and are public. This already limits the risk of leaking any private or sensitive information. As an additional safety measure, we manually verified on $\sim$~50 videos that the youtube descriptions do not contain any reference to individuals or authors; rather they contain generic links to news agency website and their social media pages. Further, we only plan to release video ids of the videos. This gives the creator the `right to be forgotten' from YouTube and from our dataset as well. These steps taken together minimize the risk of compromising privacy/identity of any individual.

\paragraph{Social Impact}

Our dataset have been sourced from news domain to ensure richness of events. However, as news media tend to be opinionated, and sometimes sensationalized, the dataset risks suffering from racial, gender, cultural and religious bias. As discussed in previous sections, we consciously took steps to reduce these biases by making an effort to collect factual news from `Center' rated news media sources (\Cref{sec:dataset}). In addition, we release the dataset sheet from \citet{Gebru2021DatasheetsFD} describing additional details aimed at limiting these biases (see \cref{appendix:dataset-sheet}). Despite these efforts, we accept that there is potential for these negative harms to creep into our model. As such, at this point, we do not recommend deployed use-cases. 

We ammortize these harms, to some extent, by showing the applicability of our method to events related to natural disasters (fire evacuation \cref{fig:task_diff}). The event hierarchy extracted from this data can be used to analyse and improve relief efforts. Further, event hierarchies can be used for validation and cross-checking to fight disinformation. We believe the potential for positive social benefits of our work outweighs the harms, and our work can be valuable to the community in the long run.

Moreover, the potential negative societal impacts are not unique to this work; rather are endemic to general machine learning models trained on publicly available data \cite{merlot-2021, miech2019howto100m, clip-2021}. There is an ongoing discourse on how to properly address these issues at a community level. We hope our usage of official public news videos and release strategy have taken a step forward in this direction. 

\begin{figure}[t]
    \centering
    \includegraphics[width=\linewidth]{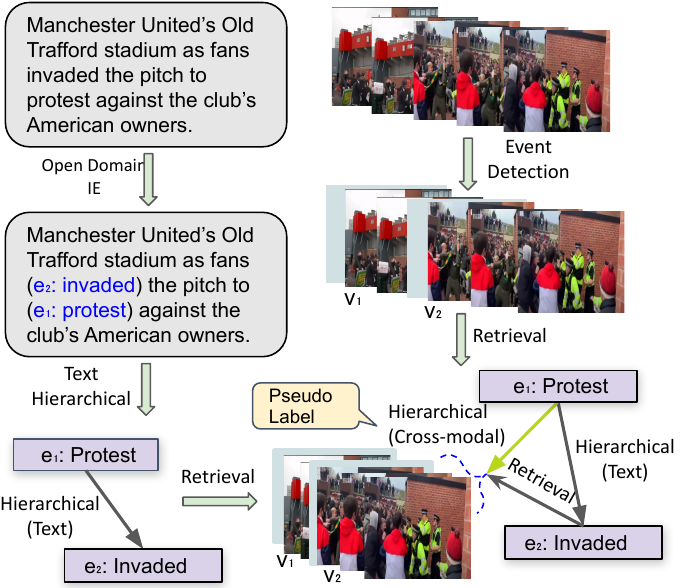}
    \caption{Proposed pseudo label generation pipeline.}
    \label{fig:pseudo_label}
\end{figure}

\section{Model Components Details}
\label{appendix:method}

\subsection{CLIP For Video Event Retrieval}
\label{appendix:pseudo-label-gen}

As discussed in the main paper, we employ a weakly supervised training method due to supervised training data constraints. As the first step in this strategy, we collect pseudo labels. Given a paired text article and video, we first extract all text events $\{e_i\}_{i=1}^m$ and all video events $\{v_j\}_{j=1}^n$. Next, we detect all hierarchical pairs in text, $\{e_ue_{u_s}\}_{u=1,s=1}^{u=p,s=q}$, using \cite{wang-etal-2021-learning-constraints}. Here, $e_u$ is the parent event and $e_{u_s}$ is the subevent. Finally, we employ CLIP to match text subevent $e_{u_s}$ with a video event so that we can propagate the hierarchical relation from text subevent to video subevent, giving us the required multimodal hierarchical relation (pseudo label) from text parent event to video subevent. We describe the precise approach for this matching step below.

CLIP is a multimodal transformer model which encodes text into an embedding using a transformer architecture, $\mathit{f}_t(.)$, and separately encodes image into an embedding, in the same embedding space as the text embedding, using another transformer architecture, $\mathit{f}_i(.)$. 
It then scores the alignment of text and image using a distance scoring function -- typically cosine -- between encoded text embedding and encoded image embedding.

While encoding text, CLIP brackets the input sentence with $\mathrm{[SOS]}$ and $\mathrm{[EOS]}$ tokens and outputs the activation of the $\mathrm{[EOS]}$ token from the highest layer of text transformer as the embedding representation for the whole sentence. However, we want the embedding of the event word contextualized by words around it. To this end, we change the attention masking of $\mathrm{[EOS]}$ token in the last layer of text transformer from being uniform to all words in the sentence to focusing on the event word with a polynomially decreasing attention mask depending on the distance of the words in sentence from the event word. We represent this modified text encoder by $\mathit{f'}_t(.)$.

We encode the text event $e_{u_s}$ by inputting the sentence containing the event word, $se_{u_s} = [w_1, w_2, ..., e_{u_s}, ...,w_i,...w_n$, to $\mathit{f'}_t$ to get $\mathit{f'}_t(se_{u_s})$. To encode the video event, $v^{u_s}_l$, we first represent it as a stack of frames sampled at some $f_s$ frames per second, $v^{u_s}_l = \{F^{u_s}_{l_y}\}_{y=1}^Z$. 
Next, we encode each frame using the image encoder, $\mathit{f}_i(.)$, and aggregate the contribution of each frame to get the video event embedding: $\frac{1}{Z}\sum_{y=1}^Z \mathit{f}_i(F^{u_s}_{l_y})$.
We say that the text event, $e_{u_s}$, represent the same event as the video event, $v^{u_s}_l$, if:
\begin{equation}
\label{eq:pseudo-label-filter}
  \mathit{f'}_t(se_{u_s}) \times \frac{1}{Z}\sum_{y=1}^Z \mathit{f}_i(F^{u_s}_{l_y}) > \lambda.  
\end{equation}
$\lambda$ is a threshold that we arrived at by fine-tuning on the multimodal event pairs which had 'Identical' relations from 100 video article sample pairs in the validation set. The threshold found, $\lambda=30.39$ gave a precision value of 75\% on the validation set. Using \cref{eq:pseudo-label-filter}, we arrive at the required text event - video event matching pairs, $\{e_{u_s}v^{u_s}_l\}_{l=1}^r$.

\subsection{Quality of Pseudo Labels}
\label{appendix:pseudo-label-quality}

The process of generating pseudo labels consists of three components - event detection, textual event-event relations detection and text to video retrieval (as shown in \Cref{fig:pseudo_label}). To evaluate the quality of pseudo labels, we examine each component individually and then look at the overall quality.

We evaluate the error propagation from incorrectly predicted textual events to downstream event-event relations prediction task by using predicted textual events, instead of ground truth text events, as input to MASHER for multimodal event relations prediction.  The system is evaluated in an end-to-end manner on the multimodal relations prediction task. We report the results in \Cref{appendix:iete2ve}. The avg. $F_1$ score for our model in this setting (called IETE2VE) is 7.8, down from 12.8.

We do not evaluate video event detection results as video event annotation can be quite ambiguous in open-vocabulary setting. For example, a video showing protest can be annotated as marching event by one annotator or as protest event by another. This difficulty has also been well noted by other works \cite{Sadhu_2021_CVPR}. In this work, we followed a coarse definition of video events (shots) which was deemed reasonable and agreed upon by the annotators.

For textual event relations detection, we used a SOTA method \cite{wang-etal-2021-learning-constraints} whose performance on a news domain benchmark \cite{glavavs2014hieve} is 52.2\% F$_1$. Since, our domain is same, we expect similar performance. Exact evaluation on our dataset is prohibitive due to the scale of annotation effort required.

For text to video retrieval, manual evaluation on a 100 samples gives a 75\% precision \Cref{appendix:pseudo-label-gen}. We do not evaluate recall because the scale of the dataset can compensate for low recall on pseudo labels.

The overall quality of pseudo labels can be judged from the performance of MM Base. in \Cref{tab:baselines-val} as pseudo label generation pipeline is the same as MM Base..

\subsection{Commonsense Features from Knowledge Base}
\label{appendix:method-train}

\begin{figure}
    \centering
    \includegraphics[width=\linewidth]{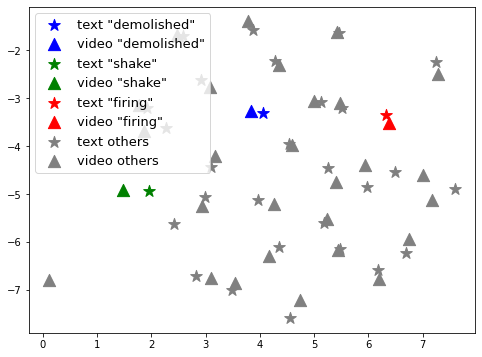}
    \caption{TSNE plot of CLIP's text event embedding and CLIP's video event embedding after passing through Contextual Transformer (CT). We see that the text and video embedding of the same event lie close together in the latent space.}
    \label{fig:cs-embeds}
\end{figure}

% Please add the following required packages to your document preamble:
% \usepackage{multirow}
% \usepackage[table,xcdraw]{xcolor}
% If you use beamer only pass "xcolor=table" option, i.e. \documentclass[xcolor=table]{beamer}
\begin{table}[t]
\centering
\setlength{\tabcolsep}{3pt}

\begin{tabular}{l lll lll@{\hspace{-5pt}} r}
\toprule
            & \multicolumn{3}{c}{Hierarchical}                                       & \multicolumn{3}{c}{Identical}                                          &                         \\
\cmidrule(lr){2-4}\cmidrule(lr){5-7}
            & \multicolumn{1}{c}{P} & \multicolumn{1}{c}{R} & \multicolumn{1}{c}{$F_1$} & \multicolumn{1}{c}{P} & \multicolumn{1}{c}{R} & \multicolumn{1}{c}{$F_1$} & \multirow{-2}{*}{Avg $F_1$} \\
\toprule

MASHER$_{\text{CT}\rightarrow CS}$  & \textbf{21.9}                 & {22.1}                  & \textbf{22.0}                     & {8.2}                   & \textbf{44.5}                 & {13.9}                   & \textbf{18.0}   \\
MASHER$_{\text{CLIP}\rightarrow CS}$ & {18.3}                  & \textbf{22.4}                  & {20.2}                     & \textbf{8.3}                   & 44.2                  & \textbf{14.0}                   & 17.1                                            \\
\bottomrule
\end{tabular}
\caption{Ablation study on the input to $CS(.,.)$. MASHER$_{\text{CT}\rightarrow CS}$ denotes the visual input to $CS$ is the output of CT (CLIP visual embedding passed through CT). MASHER$_{\text{CLIP}\rightarrow CS}$ denotes the visual input to $CS$ is directly the CLIP visual encoding.}
\label{tab:appendix-cs-input}
\end{table}

We leverage commonsense knowledge in the world to help us predict multimodal relations between open domain events. To this end, we use ConceptNet \cite{conceptnet-2017}, a large scale knowledge base containing concepts, entities and events as nodes and the relation between them as edges. We extract positive and negative event pairs from ConceptNet and use it to train the commonsense feature extractor, $CS(.,.)$. $CS(.,.)$ is a single linear layer which concatenates the embeddings of the event pairs to form its input. It outputs a 512 dimensional embedding as commonsense features, which is trained using contrastive loss. These extracted features characterize how `Hierarchical' event pairs should relate to each other. 

We first train $CS(.,.)$ offline and then use it as a black box commonsense feature extractor while training MASHER as discussed in \Cref{subsec:training}. The inputs to $CS(.,.)$ during its offline training are textual event pairs from ConceptNet. However, while using it as black box feature extractor during MASHER training, one of the input is visual while the other is textual. We argue that we can change one of the modality from textual to visual in the latter phase because we use CLIP embeddings in both phases, and CLIP's textual and visual embeddings lie close together in the same latent space.

Specifically, the inputs to $CS(.,.)$, while using it as a black box feature extractor during MASHER training, are: 1) CLIP encoding of text event, and 2) CLIP encoding of paired visual event after it has been contextualized by the Contextual Transformer (CT). We visualize in \Cref{fig:cs-embeds} that CLIP's encoding of a text event lies close to its encoding of corresponding video event which has passed through CT. It could be argued that passing CLIP encoding of visual event through CT changes the space of output embeddings. Hence, the resulting embeddings from CT can't be assumed to lie in the same embedding space as CLIP's textual event embedding, making it unsuitable as input to $CS(.,.)$. We consider this argument further by running an ablation in \Cref{tab:appendix-cs-input} by changing the input of $CS(.,.)$ from the output of CT to directly the output of CLIP visual encoding. As can be seen from the table, the Avg $F_1$ score drops from 18.0 to 17.1. This empirical result further encourages us to utilize the output of CT as input to $CS(.,.)$ in MASHER. One possible explanation of this result could be that contextualization helps improve the representation of the visual events while keeping the embedding space nearly the same.

% Please add the following required packages to your document preamble:
% \usepackage{multirow}
% \usepackage[table,xcdraw]{xcolor}
% If you use beamer only pass "xcolor=table" option, i.e. \documentclass[xcolor=table]{beamer}
\begin{table}[t]
\centering
\setlength{\tabcolsep}{5pt}

\begin{tabular}{l rrr rrr r}
\toprule
    & \multicolumn{3}{c}{Hierarchical}                                       & \multicolumn{3}{c}{Identical}                                          &                         \\
    \cmidrule(lr){2-4} \cmidrule(lr){5-7}
    & \multicolumn{1}{c}{P} & \multicolumn{1}{c}{R} & \multicolumn{1}{c}{$F_1$} & \multicolumn{1}{c}{P} & \multicolumn{1}{c}{R} & \multicolumn{1}{c}{$F_1$} & \multirow{-2}{*}{Avg $F_1$} \\
\toprule
$n=1$ & \textbf{21.9}                  & 22.1                  & \textbf{22.0}                     & \textbf{8.2}                   & 44.5                  & \textbf{13.9}                   & \textbf{18.0}                                            \\
$n=2$ & 16.5                  & 24.3                  & 19.6                   & 8.0                     & 46.4                  & 13.6                   & 16.6                                             \\
$n=4$ & 20.0                    & 19.4                  & 19.7                   & 7.9                   & \textbf{49.3}                  & 13.6                   & 16.7                                            \\
$n=6$ & 15.5                  & \textbf{31.2}                  & 20.7                   & 7.9                   & 47.8                  & 13.6                   & 17.2           \\
\bottomrule
\end{tabular}
\caption{Ablation studies on the number of layer $n$ in the contextual Transformer.}
\label{tab:ct}
\end{table}

% Please add the following required packages to your document preamble:
% \usepackage{multirow}
% \usepackage[table,xcdraw]{xcolor}
% If you use beamer only pass "xcolor=table" option, i.e. \documentclass[xcolor=table]{beamer}
\begin{table}[t]
\centering
\setlength{\tabcolsep}{4pt}

\begin{tabular}{l rrr rrr r}
\toprule
            & \multicolumn{3}{c}{Hierarchical}                                       & \multicolumn{3}{c}{Identical}                                          &                         \\
\cmidrule(lr){2-4}\cmidrule(lr){5-7}
            & \multicolumn{1}{c}{P} & \multicolumn{1}{c}{R} & \multicolumn{1}{c}{$F_1$} & \multicolumn{1}{c}{P} & \multicolumn{1}{c}{R} & \multicolumn{1}{c}{$F_1$} & \multirow{-2}{*}{Avg $F_1$} \\
\toprule
MASHER  & \textbf{21.9}                 & {22.1}                  & \textbf{22.0}                     & \textbf{8.2}                   & 44.5                  & \textbf{13.9}                   & \textbf{18.0}   \\
% MERP w/ WS  & 14.9               & 26.4                & 19.1                   & 8.0           & \textbf{48.9}                 & 13.7             & 16.4 \\
\hspace{2pt} - w/ MIL & 7.4   & \textbf{47.7}     & 12.8          & 7.3  & 47.5                  & 12.7 & 12.8  \\
\bottomrule
\end{tabular}
\caption{Effect of multiple instance learning (MIL) strategy}
\label{tab:identical-mil}
\end{table}

% We then use these structures as supervision to train a multimodal classifier directly on the multimodal event relation recognition task. In order to further transfer knowledge about how events tend to relate to one another, we also provide our model with commonsense knowledge features as an additional input.
\subsection{Implementation Details}
\label{appendix:impl-details}

We provide additional implementation details in this section.

\begin{itemize}
    \item To extract video event features, we sample frames uniformly from the video event clip at a rate of $f_s$. $f_s$ in this work was fixed at 3 frames per second.
    \item The video event embeddings extracted using CLIP forms the token embedding to be fed to the Contextual Transformer (CT). These embeddings are concatenated by positional embeddings, as is common with transformer architectures (\cite{vaswani2017attention}). Further, we cap the maximum number of video events that CT can process at 77 to limit the computational cost.
    \item As CLIP has demonstrated strong perormance in co-relating multimodal data, we use CLIP to remove false positive `Identical' relations from model prediction. To implement it, we use \cref{eq:pseudo-label-filter} with a threshold of 28.0. Event pairs scoring below this threshold are pruned. This threshold was again arrived at using the validation set.
\end{itemize}

\section{Evaluation Metric and Baselines}
\label{appendix:exp-details}
\subsection{Mathematical Formulation For Evaluation Metric}
\label{appendix:eval-metric}

Given ground truth event-event relations $\{r_i\}_{i=1}^K$ and predicted relations $\{p_j\}_{j=1}^K`$, where $r_i$, $p_j$ $\in$ \{`Hierarchical', `Identical', `NoRel'\}, we define Precsiion ($P_t$), Recall ($R_t$) and $F_1$ ($F_{1_t}$) for a relation, $t \in$ \{`Hierarchical', `Identical'\}, as:

\[
P_t = \frac{|\{p_i, \forall p_i=t\} \cap \{r_i, \forall r_i=t\}|}{|\{p_i, \forall p_i=t\}|} \]
\[
\quad R_t = \frac{|\{p_i, \forall p_i=t\} \cap \{r_i, \forall r_i=t\}|}{|\{r_i, \forall r_i=t\}|} \]
\[
\quad F_{1_t} = \frac{P_tR_t}{P_t+R_t}
\]

Then, Avg $F_1 = \frac{\sum_tF_{1_t}}{2}$.
\subsection{Text Only Baseline Details}
\label{appendix:baselines}

We construct text only baseline by using video ASR as proxy for video itself -- each video event is represented by the ASR content within the timestamps of the video event. Now, events and event-event relations with text article can be extracted from ASR using NLP models. However, we don't use raw ASR directly. We found it difficult because they are unpunctuated and the NLP models to detect events and event-event relations have been trained on punctuated data. As such, we punctuate ASR using \cite{bert-restore-punctuation}.

We observed the Text baseline to have a low recall in \Cref{tab:appendix-network-arch}. On manual analysis, we found three reasons for it: 1) a number of video events do not associated ASR content; 2) some video events are very short in duration which limits the ASR content to a very few words; and 3) text events detected in ASR sometimes don't provide useful information about the events in video (eg. an event such as `mentioned' detected in ASR).

% Please add the following required packages to your document preamble:
% \usepackage{multirow}
% \usepackage[table,xcdraw]{xcolor}
% If you use beamer only pass "xcolor=table" option, i.e. \documentclass[xcolor=table]{beamer}
\begin{table}[t]
\centering
\setlength{\tabcolsep}{5pt}
\small

\begin{tabular}{l ccc ccc c}
\toprule
                                        & \multicolumn{3}{c}{Hierarchical}                                       & \multicolumn{3}{c}{Identical}                                          &   \multirow{2}{*}{Avg $F_1$}                       \\
                     \cmidrule(lr){2-4} \cmidrule(lr){5-7}   
                                        & P & R & $F_1$ & P & R & $F_1$ &  \\ 
\toprule
Prior Base.                  & 4.7                  & 4.7                & 4.7                    & 2.0                  & 2.0                 & 2.0                    & 3.4                                            \\
Text Base.               & 5.9                  & 0.1                  & 0.1                   & 2.5                  & 7.1                 & 3.6                    & 1.9                                             \\
MM Base. & \textbf{35.7}                 & 5.0                  & 8.8                    & \textbf{8.8}                  & 33.1                 & \textbf{13.9}                   & 11.4 \\
Video-LLaMA & 4.8 & 13.15 & 7.06 & 1.76 & 4.08 & 2.46 & 4.76\\
BLIP-2 & 0.0                & 0.0                  & 0.0                   & 27.08                 & 5.68                & 9.39                   & 4.7 \\ \midrule
MASHER &   21.9 & \textbf{22.1} &	\textbf{22.0} &	8.2	& \textbf{44.5}	& \textbf{13.9}	& \textbf{18.0}  \\   
\bottomrule
\end{tabular}
\caption{Comparison with baseline models on the validation set.}
\label{tab:appendix-blip}
\end{table}

% Please add the following required packages to your document preamble:
% \usepackage{multirow}
% \usepackage[table,xcdraw]{xcolor}
% If you use beamer only pass "xcolor=table" option, i.e. \documentclass[xcolor=table]{beamer}
\begin{table}[t]
\centering
\setlength{\tabcolsep}{3pt}

\begin{tabular}{l rrr rrr r}
\toprule
            & \multicolumn{3}{c}{Hierarchical}                                       & \multicolumn{3}{c}{Identical}                                          &                         \\
\cmidrule(lr){2-4}\cmidrule(lr){5-7}
            & \multicolumn{1}{c}{P} & \multicolumn{1}{c}{R} & \multicolumn{1}{c}{$F_1$} & \multicolumn{1}{c}{P} & \multicolumn{1}{c}{R} & \multicolumn{1}{c}{$F_1$} & \multirow{-2}{*}{Avg $F_1$} \\
\toprule

MASHER  & \textbf{21.9}                 & \textbf{22.1}                  & \textbf{22.0}                     & \textbf{8.2}                   & 44.5                  & \textbf{13.9}                   & \textbf{18.0}   \\
\hspace{3pt} - w/o pruning & \textbf{21.9}                  & \textbf{22.1}                  & \textbf{22.0}                     & 7.7                   & \textbf{45.8}                  & 13.2                   & 17.6                                             \\
\bottomrule
\end{tabular}
\caption{Ablation studies on inference time ensemble with CLIP.}
\label{tab:identical-pruning}
\end{table}
% Please add the following required packages to your document preamble:
% \usepackage{multirow}
% \usepackage[table,xcdraw]{xcolor}
% If you use beamer only pass "xcolor=table" option, i.e. \documentclass[xcolor=table]{beamer}
\begin{table*}[t]
\centering
\small

\begin{tabular}{l ccc ccc c}
\toprule
                                        & \multicolumn{3}{c}{Hierarchical}                                       & \multicolumn{3}{c}{Identical}                                          &   \multirow{2}{*}{Avg $F_1$}                       \\
                     \cmidrule(lr){2-4} \cmidrule(lr){5-7}   
                                        & P & R & $F_1$ & P & R & $F_1$ &  \\ 
\toprule
Prior Base.                  & 0.9/0.2                  & 0.2/0.2                 & 0.2/0.2                    & 0.1/0.1                  & 0.1/0.1                 & 0.1/0.1                    & 0.1/0.1                                             \\
Text Base.               & 3.9/0.0                  & 0.1/0.0                  & 0.1/0.0                    & 0.5/1.6                  & 3.6/10.0                 & 0.8/2.8                    & 0.5/1.4                                             \\
MM Base. & \textbf{21.8}/\textbf{11.2}                 & 2.2/1.7                  & 3.9/2.9                   & \textbf{5.0}/\textbf{6.7}                  & 23.9/26.1                 & \textbf{8.2}/\textbf{10.7}                   & 6.1/6.8                                          \\ \midrule
MASHER &   5.5/5.6 & \textbf{6.4}/\textbf{7.6} &	\textbf{5.9}/\textbf{6.5} &	4.2/5.3	& \textbf{30.1}/\textbf{30.0}	& 7.3/9.0	& \textbf{6.6}/\textbf{7.8}  \\   
\bottomrule
\end{tabular}
\caption{Comparison with baseline models on the validation/test set for the IETe2Ve setting.}
\label{tab:baselines-ie}
\end{table*}

\subsection{Image-based LLM Baseline: BLIP-2}
We consider a large language image-based multimodal baseline: BLIP-2 \cite{li2023blip2}. This was done to study how image-based LLM can generalize to the proposed video task. As BLIP-2 is image based, not video based, we separately input 5 uniformly sampled frames from the video event and take the majority prediction as output. Along with the frame we input text event and prompt it with the definition of the task and the three types of relations. We report the results in \Cref{tab:appendix-blip}. BLIP-2 fails to predict any relation as Hierarchical. This is not surprising as this is a completely new task. However, BLIP-2 does perform better than Text Base. for Identical relations, demonstrating the value of visual data to this task. Indeed, it's telling that it performs worse than MM Base. Perhaps, it indicates the superiority of CLIP in multimodal matching tasks.

\section{Ablation and Hyperparameter Details}
\label{appendix:ablation-study}

\paragraph{Network Architecture Ablation} We have reported the ablation study on the components of our mdoel, MASHER, in \Cref{tab:appendix-network-arch}. We observe that the contribution of contextual transformer is most significant towards the full model performance. This makes sense because contextualizing a video event with respect to other video event can significantly change its interpretation. For example, a video event showing crowd of people on streets could be demonstrators agitating against the government or cheering on their newly elected leader depending on the context.

\paragraph{Effectiveness of CLIP ensemble during Inference}
We ablate the effect of inference time ensemble with CLIP in \Cref{tab:identical-pruning}. As expected, it improves performance by eliminating identical false positives.

\paragraph{Influence of Number of Layers in Contextual Transformer.} To further investigate the influence of different number of layer $n$ in the contextual Transformer, we conducted a set of ablations by setting $n=\{1, 2, 4, 6\}$, and results are reported in Table~\ref{tab:ct}. As can be observed, all different layer settings can achieve good Avg $F_1$ score, which demonstrates robustness of our model. Meanwhile, $n=1$ achieves slightly better results. The reason could be that more layers cause some over-fitting. As such, we use $n=1$ in setting in our best model.

% To this end, we used $n=1$ is all setting.

\paragraph{Dealing with Noisy Labels} As discussed in the main paper, our training labels are actually pseudo labels derived using other NLP and Vision models. As such, the labels are naturally noisy. To deal with this noise, we employed Multi Instance Learning (MIL) \cite{mil_2018} training objective (\Cref{tab:identical-mil}). However, we found that it didn't improve performance. One possible reason could be the fact that our large scale training data compensates for the inherent noise in labels. Such outcome have also been observed previously by \cite{Jia2021ScalingUV}.

\section{Qualitative Analysis on predicted samples}
\label{appendix:qual-analysis}

\begin{figure*}[]
    \centering
        \includegraphics[width=\textwidth]{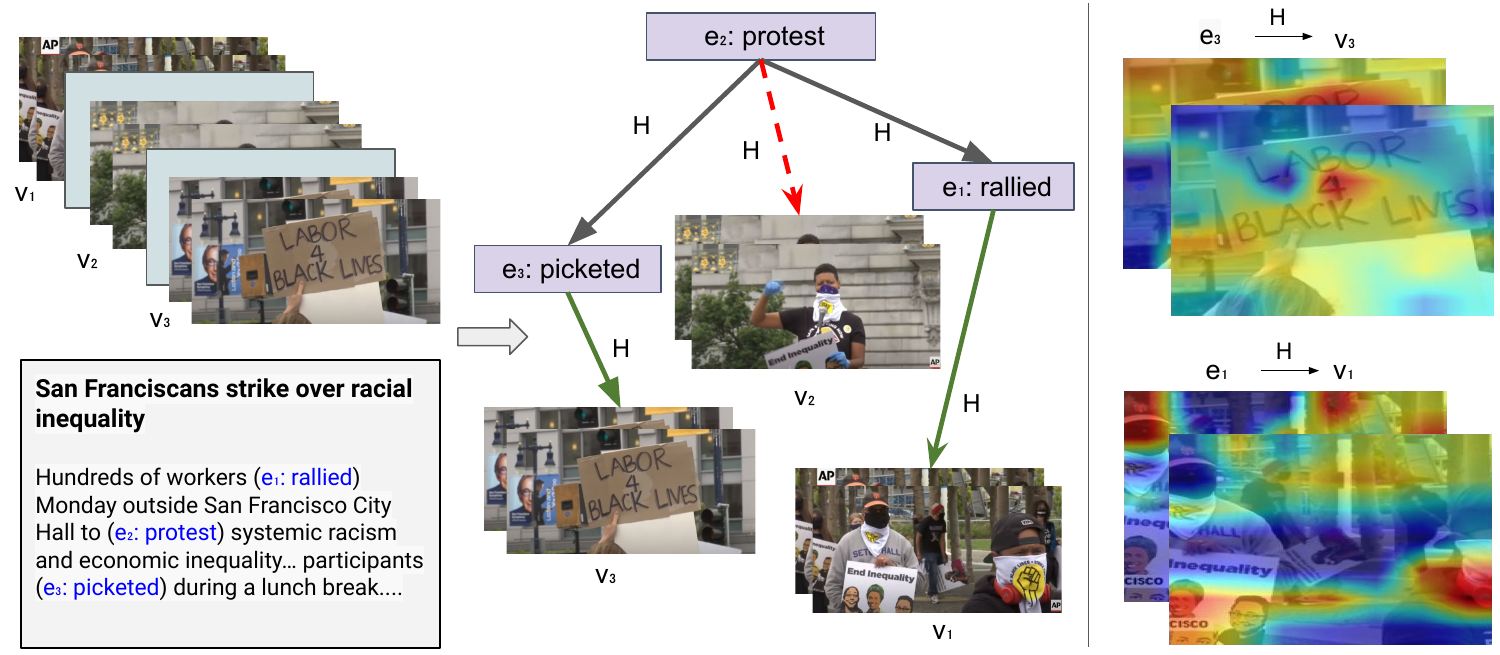}
         \includegraphics[width=\textwidth]{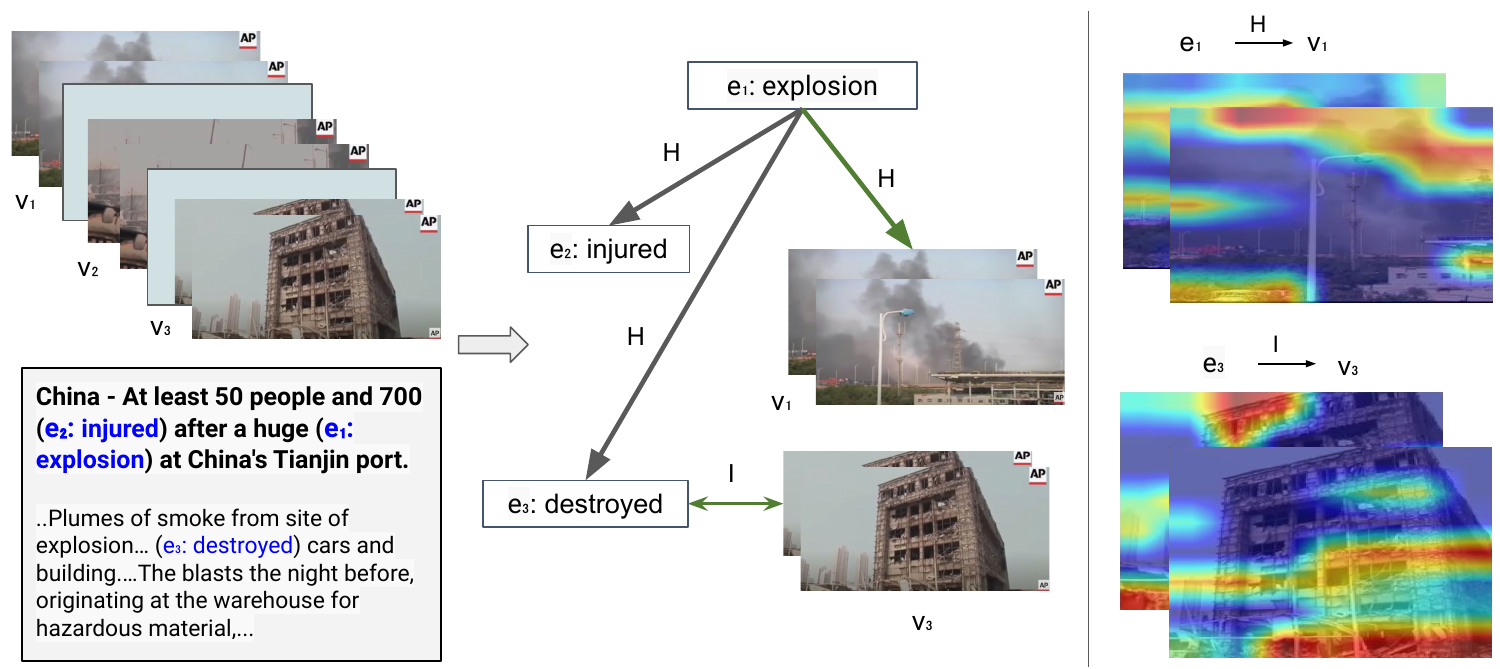}
         \includegraphics[width=\textwidth]{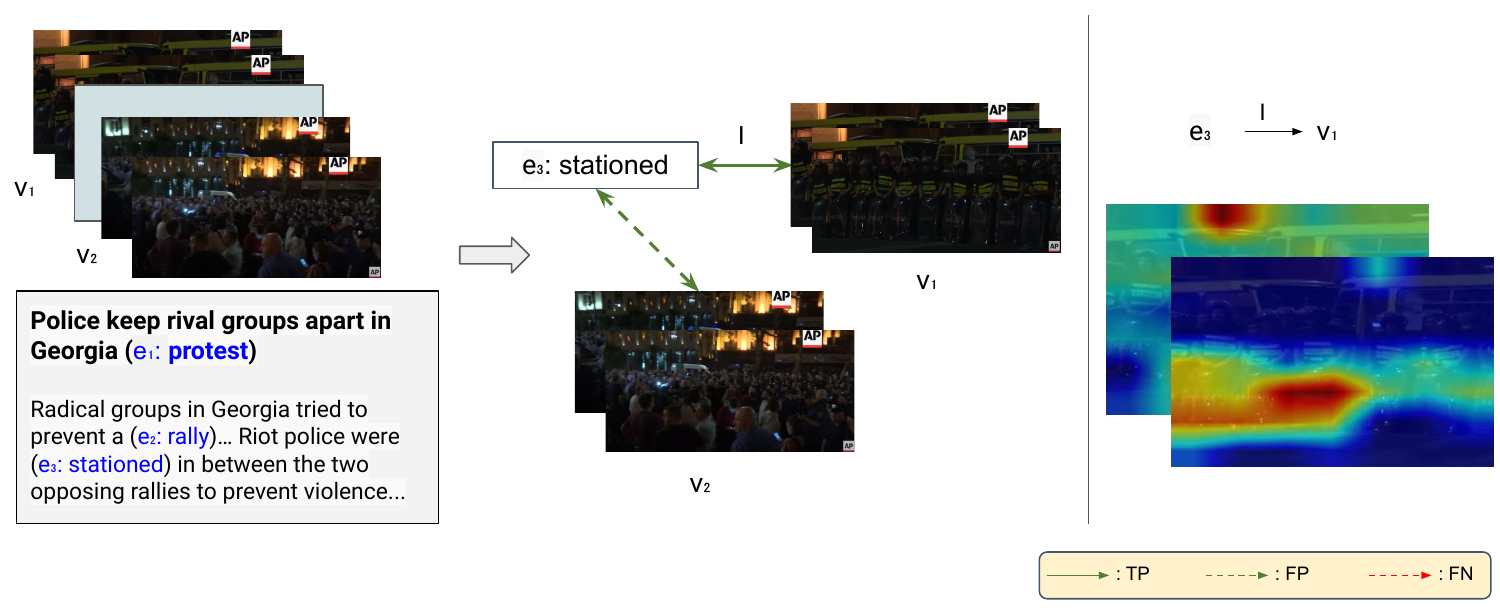}
     \caption{Model Prediction: The left column shows the input, middle column shows the model output and the right column shows heatmap visualizations of the input based on the prediction.}
    \label{fig:appendix-qual-samp}
\end{figure*}

We provide additional qualitative results and visualizations in \Cref{fig:appendix-qual-samp}. As can be seen, our model is able to enrich text event graph with new visual events nodes and different multimodal relations from diverse topics (explosion and protest). This is especially emphatic because our model was trained in an open domain setting and was not specifically trained to detect events from these topics.

Further, we compute heatmap visualization for illustrating the attention of the model while making predictions. We use Grad-CAM for this purpose \cite{gradcam, jacobgilpytorchcam}. Through these visualizations, we provide evidence that the model is indeed learning meaningful co-relations between text and visual events instead of learning some statistical prior. For example, in top row we can see that the model is focussing on smoke while predicting it's hierarchically related to explosion.

\section{End-to-End System Evaluation: Event Detection + Relationship Prediction}
\label{appendix:iete2ve}

Multimodal Event graphs consists of text and video events as nodes and the event-event relations relations as edges. The proposed task in the paper, focuses on prediction and evaluation of multimodal event-event relations, given a text and video event. The assumption is that we are provided with ground truth text and video events. However, the assumption doesn't hold for `in the wild' text-video pair. To generate multimodal event graphs for such an input, we need to predict text and video events first and then predict the relations between them. As such, we evaluate our model against this combined task as well -- event detection and event-event relation prediction. 

In this setting, the model is first required to predict text event and then predict multimodal event-event relation between the predicted text event and given video event. We don't task the model to predict video events because our video events are actually camera shots by definition and are not annotated by the annotators. The evaluation is then done between the predicted multimodal event-event relations and ground truth multimodal event-event relations. The evaluation metric remains the same -- Precision, Recall and $F_1$. We call this setting IETe2Ve (Information Extraction + Text event to Video event relation prediction) and the earlier setting of predicting only multimodal event-event relations Te2Ve(Text event to Video event relation prediction). 

In our model, we use \cite{shen-etal-2021-corpus} to extract text events and then predict the multimodal relations the usual way. For Text Base. and MM Base. as well, we again use \cite{shen-etal-2021-corpus} as the first step to detect text events. The results are reported in \Cref{tab:baselines-ie}. MASHER obviously performs worse now than in Te2Ve setting (7.8 vs 12.8 on test set). However, it still performs better than next best baseline by 1\%.

We note the extreme challenge posed by this setting -- predicting open domain events and then predicting relations between it and open domain video events. This is also evidenced by extremely low performance of Prior Base. -- 0.1 Avg $F_1$ down from 1.62 in Te2Ve setting. This is so because in IETe2Ve, it must consider all words in text article as possible events. This makes the prior of `NoRel' very high, significantly reducing number of relations predicted as `Hierarchical' and `Identical'. We also note the low performance of Text Base. -- this is due to the same reasons discussed in \Cref{appendix:eval-metric}.

\section{Limitations and Future Directions}
\label{appendix:limitations}

There are several limitations of our work which can be studied in future work. First, we have explored two types of event relations, hierarchical and identical, but additional relation types like temporal and causal can be studied. Also, multimodal relations directed from video to text can also be discovered. Finally, utilizing a prior grounding technique in the pseudo label generation pipeline leads to noisy pseudo labeling of hierarchical relations as identical resulting in poor performance. Better pseudo label generation techniques, like graph alignment between text and video hierarchies, can be explored in future.

% \todo{Add dataset sheet}

\section{Datasheet for \datasetname}
\label{appendix:dataset-sheet}
% \label{supp:datasheets}

In this section, we present a DataSheet \cite{gebru2018datasheets, bender2018data} for \datasetname, synthesizing many of the other analyses we performed in this paper.

\begin{enumerate}

\item Motivation For Datasheet Creation
\begin{itemize}
  \item \textbf{Why was the dataset created?} 
  In order to study cross-modal event event relations.
  \item \textbf{Has the dataset been used already?}
  No.
\item \textbf{What (other) tasks could the dataset be used for?}
Possibly visual grounding and multi-modal schema induction.

% \item \textbf{Who funded dataset creation?}
% This work was funded by DARPA MCS program through NIWC Pacific (N66001-19-2-4031), and the Allen Institute for AI.
\end{itemize}

\item Data composition
\begin{itemize}
\item \textbf{What are the instances?}
The instances that we consider in this work are pairs of news articles and associated videos .

\item \textbf{How many instances are there?}
% We include 6 million videos. The total length of all the ASR transcripts is 5 billion BPE tokens. Altogether, we extracted 180 million image frames from this data.
We include 100.5K news articles and associated videos.

\item \textbf{What data does each instance consist of?}
% The instances have `raw' video frames and text, which we preprocess through BPE tokenization and extracting frames for every 32 BPE tokens.
The instances have `raw' video frames and text.

\item \textbf{Is there a label or target associated with each instance?}
% We only use the ASR captions as labels in this work, though it might be also possible to use auxiliary information (like tags or video titles).
The train set doesn't contain any label, while the test set is annotated with text events and their multimodal relations (`Hierarchical' or `Identical') with video events.

\item \textbf{Is any information missing from individual instances?}
No.

\item \textbf{Are relationships between individual instances made explicit?}
Not applicable -- we do not study relations between different videos and articles.

\item \textbf{Does the dataset contain all possible instances or is it a sample?}
Just a sample.

\item \textbf{Are there recommended data splits (e.g., training, development/validation, testing)?}
% We do not provide recommended data splits at this time, as this data was built only for pretraining rather than evaluation. We suspect that the data is large enough that overfitting is not a major concern.
We divide our data into train/val/test splits.

\item \textbf{Are there any errors, sources of noise, or redundancies in the dataset? If so, please provide a description.}
Yes. The Inter Annotator Agreement scores for Hierarchical multimodal relations is 47.5 which denotes inconsistencies in the annotation of different annotators. However, we argue that this task is difficult and even in NLP, where this task is unimodal, the IAA scores of two established datasets \cite{glavavs2014hieve, hovy-etal-2013-events} are 69 and 62 respectively.
% \textcolor{red}{TBD}

\item \textbf{Is the dataset self-contained, or does it link to or otherwise rely on external resources (e.g., websites, tweets, other datasets)?}
The dataset is self-contained. However, we only release video urls for the sake of minimizing privacy and consent issues.
% \textcolor{red}{TBD}

\end{itemize}
\item Collection Process
\begin{itemize}\item \textbf{What mechanisms or procedures were used to collect the data?}
We used the pytube \footnote{\url{https://pytube.io/en/latest/}} and youtube-search-python \footnote{\url{https://pypi.org/project/youtube-search-python/}} library. 
% \textcolor{red}{TBD}

\item \textbf{How was the data associated with each instance acquired? Was the data directly observable (e.g., raw text, movie ratings), reported by subjects (e.g., survey responses), or indirectly inferred/derived from other data?}
The data was directly observable from YouTube.
% \textcolor{red}{TBD}

\item \textbf{If the dataset is a sample from a larger set, what was the sampling strategy (e.g., deterministic, probabilistic with specific sampling probabilities)?}
We collected random 100.5K youtube video and description pairs from 9 media sources rated Center.
% \textcolor{red}{TBD}

\item \textbf{Who was involved in the data collection process (e.g., students, crowdworkers, contractors) and how were they compensated (e.g., how much were crowdworkers paid)?}
Data collection was primarily done by the first author of this paper.
% \textcolor{red}{TBD}

\item \textbf{Over what timeframe was the data collected? Does this timeframe match the creation timeframe of the data associated with the instances (e.g., recent crawl of old news articles)? If not, please describe the timeframe in which the data associated with the instances was created.}
The data was collected from August 2021 to September 2021. Although, the data was created by the news media sources as early as 1963. They were uploaded on the YouTube platform since it was established.
% \textcolor{red}{TBD}
\end{itemize}

\item {Data Preprocessing}
\begin{itemize}\item \textbf{Was any preprocessing/cleaning/labeling of the data done (e.g., discretization or bucketing, tokenization, part-of-speech tagging, SIFT feature extraction, removal of instances, processing of missing values)?}
% Yes, we discuss this in Appendix~\ref{sec:scraping}: of note, we use a sequence-to-sequence model to `denoise' ASR transcripts (Appendix \ref{ssec:denoise}), BPE-tokenize text, turn everything into segments, and extract the middle image frame for each video segment.
We do not preprocess or label the train set. However, we do label the test set for multimodal event-event relations.
% \textcolor{red}{TBD}

\item \textbf{Was the “raw” data saved in addition to the preprocessed/cleaned/labeled data (e.g., to support unanticipated future uses)? If so, please provide a link or other access point to the `raw' data.}
The raw data was saved, but at this time we do not plan to release it directly due to copyright and privacy concerns.

\item \textbf{Is the software used to preprocess/clean/label the instances available? If so, please provide a link or other access point.}
% We will make our code public to support future research.
We will release our code upon the acceptance of publication.

\item \textbf{Does this dataset collection/processing procedure achieve the motivation for creating the dataset stated in the first section of this datasheet? If not, what are the limitations?}
% We believe our dataset does allow for study of our goal -- indeed, it covers grounded temporal situations from a variety of domains -- but with significant limitations. Some of the key ones we are aware of involve various biases on YouTube, which we discuss in Section~\ref{sec:sec_with_potential_impacts}.
The qualitative results (\Cref{fig:appendix-qual-samp}) and data exploration (Ccref{fig:data-wordle}) illustrate rich diversity of events and their relations, implying our goal of creating a dataset for studying multimodal events relations was successful. However, there are limitations which we discuss in \Cref{appendix:dataset-licensing}.
% \textcolor{red}{TBD}
\end{itemize}
\item Dataset Distribution
\begin{itemize}\item \textbf{How will the dataset be distributed?}
% At this time, we plan to distribute all the metadata (transcripts, etc) that we used, as well as links to the YouTube videos that we used. We will do this on our website.
We plan to distribute all the videos and articles.

\item \textbf{When will the dataset be released/first distributed? What license (if any) is it distributed under?}
% We will release it as soon as possible, using a permissible license for research-based use.
We will release the dataset upon acceptance of publication.

\item \textbf{Are there any copyrights on the data?}
% We believe our use is `fair use,' however, due to an abundance of caution, we will not be releasing any of the videos themselves.
It should be ``fair use''.

\item \textbf{Are there any fees or access restrictions?}
No.

\end{itemize}
\item Dataset Maintenance
\begin{itemize}
%     \item \textbf{Who is supporting/hosting/maintaining the dataset?}
% The first authors of this work.

\item \textbf{Will the dataset be updated? If so, how often and by whom?}
% We do not plan to update it at this time.
No.

\item \textbf{Is there a repository to link to any/all papers/systems that use this dataset?}
% Not right now, but we encourage anyone who uses the dataset to cite our paper so it can be easily found.
No.

\item \textbf{If others want to extend/augment/build on this dataset, is there a mechanism for them to do so?}
Not at this time.
\end{itemize}

\item Legal and Ethical Considerations
\begin{itemize}

\item \textbf{Were any ethical review processes conducted (e.g., by an institutional review board)?}
% No official processes were done, as our research is not on human subjects, but we had significant internal deliberation when choosing the scraping strategy.
We haven't done official processes. 

\item \textbf{Does the dataset contain data that might be considered confidential?}
No, we only use public videos.

\item \textbf{Does the dataset contain data that, if viewed directly, might be offensive, insulting, threatening, or might otherwise cause anxiety? If so, please describe why.}
% Yes -- many of these videos exist on YouTube; we discuss this more in Section~\ref{sec:sec_with_potential_impacts}.
% \textcolor{red}{TBD}
Yes, we discuss this in \Cref{appendix:dataset-licensing}.

\item \textbf{Does the dataset relate to people?}
Yes.

\item \textbf{Does the dataset identify any subpopulations (e.g., by age, gender)?}
Not explicitly.

\item \textbf{Is it possible to identify individuals (i.e., one or more natural persons), either directly or indirectly (i.e., in combination with other data) from the dataset?}
% Yes, our data includes celebrities, or other YouTube-famous people. All of the videos that we use are of publicly available data, following the Terms of Service that users agreed to when uploading to YouTube.
Yes. All the videos we use are publicly available,
\end{itemize}
\end{enumerate}

\end{document}